%% file: main.tex
\documentclass[entropy,article,preprint,pdftex,moreauthors]{Definitions/mdpi} 

\firstpage{1} 
\pubvolume{1}
\issuenum{1}
\articlenumber{0}
\pubyear{2024}
\copyrightyear{2024}
\datereceived{ } 
\daterevised{ } 
\dateaccepted{ } 
\datepublished{ } 
\hreflink{https://doi.org/} 

\usepackage[british]{babel}

\usepackage{natbib} 
    
\usepackage{mathtools} 
\usepackage{booktabs} 
\usepackage{tikz} 
\usepackage{tabularray}
\usepackage{pdflscape}

\renewcommand{\paragraph}[1]{\textbf{#1}~}

\usepackage{xcolor}

\let\oldcite\cite
\renewcommand{\cite}[1]{\ifblank{#1}{}{\oldcite{#1}}}
\makeatletter
\newcommand{\xxx}{\@ifnextchar\bgroup{\xxx@arg}{\xxx@noarg}}
\newcommand{\xxx@noarg}{\colorbox{red}{XXX}}
\newcommand{\xxx@arg}[1]{\colorbox{red}{XXX: #1}}
\makeatother

\Title{Causal knowledge engineering: A case study from COVID-19}

\TitleCitation{Causal knowledge engineering: A case study from COVID-19}

\Author{Steven Mascaro $^{1}$\orcidA{}, Yue Wu $^{2}$\orcidB{}, Ross Pearson $^{1}$\orcidC{}, Owen Woodberry $^{1}$, Jessica Ramsay $^{3}$\orcidE{}, Tom Snelling $^{2,3}$\orcidF{}, Ann E. Nicholson $^{1}$\orcidG{}}

\address{%
$^{1}$ \quad Faculty of Information Technology, Monash University, Clayton, Melbourne, Australia\\
$^{2}$ \quad Sydney School of Public Health, Faculty of Medicine and Health, University of Sydney, Sydney, Australia\\
$^{3}$ \quad Wesfarmers Centre of Vaccines and Infectious Diseases, Telethon Kids Institute, Perth, Australia}

\corres{Correspondence: steven.mascaro@monash.edu}

\abstract{COVID-19 appeared abruptly in early 2020, requiring a rapid response amid a context of great uncertainty. Good quality data and knowledge was initially lacking, and many early models had to be developed with causal assumptions and estimations built in to supplement limited data, often with no reliable approach for identifying, validating and documenting these causal assumptions. Our team embarked on a knowledge engineering process to develop a causal knowledge base consisting of several causal BNs for diverse aspects of COVID-19. The unique challenges of the setting lead to experiments with the elicitation approach, and what emerged was a knowledge engineering method we call Causal Knowledge Engineering (CKE). The CKE provides a structured approach for building a causal knowledge base that can support the development of a variety of application-specific models. Here we describe the CKE method, and use our COVID-19 work as a case study to provide a detailed discussion and analysis of the method.}


\begin{document}

\section{Introduction}\label{sec:intro}


The COVID-19 pandemic erupted abruptly in early 2020, requiring a rapid response amid a context of great uncertainty. There was a important need to better understand the disease via a diverse array of modelling approaches~\citep{shao+2020,kucharski+2020,aljame+2020,parr+2021}; many of these models had to be developed with many expert assumptions and estimations built in to supplement the limited data. In the absence of other similar work (as well as in the absence of available data), we embarked on a knowledge engineering process to elicit and construct a causal knowledge base built on Bayesian networks (BNs)~\citep{mascaro+2023,wu+2021}. The causal knowledge base was principally developed between March and July of 2020 and captured evolving knowledge and hypotheses around COVID-19 pathophysiology, testing and diagnosis.

The knowledge engineering process for this work had to come together quickly and worked within the context of the unique circumstances that existed at the time, such as social isolation, home-based work, and large volumes of novel and changing COVID-19 information; in addition, there was the expectation that large amounts of high quality clinical data would become available. We made use of existing knowledge engineering techniques~\citep{korb&nicholson2010a}, but the challenging circumstances led to our also adopting some novel approaches and extensions. We have collected these together into a general knowledge engineering method, which we call {\em Causal Knowledge Engineering} (CKE). CKE is a process for capturing qualitative causal knowledge as a set of BNs (and associated materials such as variable dictionaries) that form a causal knowledge base to support the development of application-specific BNs, as well as to support other kinds of applied models and applications. We illustrate it here using our COVID-19 experience as an example.

Fundamentally, models simulate a system to allow inference about that system. As a type of mathematical model, a BN (whether causal or not) is a directed acyclic graph (DAG) in which each node represents a random variable, and each directed arc represents a statistical relationship between variables (specifically, from a {\em parent} to a {\em child} node)~\citep{pearl1988a}. Traditionally, statistical relationships in a BN are represented by conditional probability tables (CPTs), each of which maps (i.e., is a function of) each combination of parent node states to a probability distribution over the child node. However, discrete tables are not required, and any probabilistic function of the parents (discrete or continuous) is possible, even if not all such functions are computationally convenient. In a {\em causal} BN, arcs represent causal relationships, and a causal BN is intended to {\em represent some causal process in the real world}.\footnote{Within reason: 1) an occasional convenient correlational arc is fine, so long as all required causal queries are still valid; 2) mistakes, and errors of omission and commission are always possible, and do not necessarily invalidate the intention.} An unparameterised BN in the form of a DAG can also be useful when the arcs are taken to represent probabilistic causal relationships, just as they would in a parameterised causal BN. These are commonly simply called DAGs, although we prefer {\em causal} DAGs, and they have become widely used in epidemiology, primarily for the (more restricted) purpose of causal inference~\cite{hernan&robins2020}. Here, for clarity and simplicity, we will use the term `CKBN' (`\underline{C}ausal \underline{K}nowledge base \underline{BN}') to refer to any causal BN that belongs to the causal knowledge base, whether it does or does not have parameters, whether the parameters are intended to be discrete or continuous, exact or approximate, and whatever the provenance, development or validity of the causal BN.

Knowledge engineering~\citep{studer+1998} refers to a collection of activities used to construct a {\em knowledge base} or more generally a {\em knowledge based system}. A knowledge base is simply a collection of information (i.e., meaningful data about the world) stored in a structured form and in which the relationships between pieces of information are specified, so that inferences can be performed. A (causal) BN of any kind is an example of a (causal) knowledge base because it captures information in the form of variables and their (causal) relationships. A BN coupled with an inference engine forms a knowledge based system; since inference engines for BNs are standardised and universal, most BN-based work tends to focus just on the knowledge base itself. Of course, a causal knowledge base can also consist of multiple CKBNs --- potentially containing complementary, contrasting or even contradictory information.\footnote{In order for a BN, or anything else, to be a model of some system, it should be homomorphic to that system --- that is, there should exist a function that maps constants, functions and relations from the model to that system --- as described in \citet{korb&mascaro2009}. By contrast, a knowledge base is more general and may consist of any information relating to the system, be it models, taxonomies or simple item lists.}

Knowledge Engineering with Bayesian Networks (KEBN) is a methodology proposed by \citet{korb&nicholson2010a} that specifies how knowledge engineering can be done start-to-end to construct a (typically causal) BN. KEBN generally focuses on BNs that are intended for deployment, however to develop CKE, we extracted some of the principles underlying the KEBN approach (in particular, the spiral model, along with the techniques and checks used during structure development) and extended them to provide a more detailed account of the development of a causal knowledge base with many CKBNs. In principle, however, CKE comprises the earliest (conceptual modelling) stages of the KEBN approach. It also has much in common with some earlier approaches aimed at building a conceptual model or ontology of concepts first~\citep{helsper&vandergaag2002,nadkarni&shenoy2004,fenz2012}, but aims for stricter adherence to causality and causal BN semantics, and additionally recommends the use of qualitative causal annotations as well as what we call qualitative parameterisation.

The version of CKE that we used for COVID-19 makes heavy use of knowledge elicitation supported by literature review in order to build up the causal knowledge base. However, the CKE process also involves other important activities, such as developing variable dictionaries, consulting available raw data, designing and refining the key CKBN structures based on modelling considerations, making use of qualitative parameterisations, co-ordinating and educating experts, as well as sense-checking and validation with experts. As with KEBN, CKE, including the version we used for COVID-19, involves a great deal of iteration. The CKE process is described in detail in Section~\ref{sec:ke}. We also describe methodological considerations for elicitation separately in Section~\ref{sec:elicit} owing to its importance and its prominence in the COVID-19 case study.

\subsection{COVID-19 case study}

The CKE process for our case study took place over 4 months, from March to July 2020. Due to the haste with which our team switched to the COVID-19 modelling effort, we did not yet have a specific application for the BN models in mind. This required us to consider what knowledge might be useful across a broad range of applications. We quickly excluded COVID-19 epidemiology from our scope, as a lot of work was already being done in that area (including some BN work) (e.g., \citealp{shao+2020,kucharski+2020,tang+2020a}). We instead chose 3 aspects of COVID-19 to consider: diagnosis and testing~\citep{wu+2021}; the initial pathophysiology of infection; and complications arising from severe infection~\citep{mascaro+2023}.

Despite being parameterisable, the COVID-19 causal knowledge base was not expected to ever be used directly in a deployed application —-- it was created primarily to support future models that would be application (and data) specific, allowing those models to contain just those features of the causal knowledge base that are required for the application. We will call these later models {\em application} models (or application BNs), to distinguish them from the knowledge base models.

\subsubsection{Challenges of the Case Study}

COVID-19 created a unique context for our modelling work, as well as for the modelling problem itself. This gave rise to several challenges which are described below.

\paragraph{Modelling a highly uncertain problem domain with dynamically evolving knowledge.} Very little was known about the SARS-CoV-2 virus when COVID-19 first began to create global disruption. There was little reliable data or available evidence about the behaviour of the new virus or how it presents in patients, and very little clinical experience with the disease. Global concern led to an unprecedented effort to collect data and perform research that rapidly swelled the information available on the disease. Not all of this information was compatible and reliable~\citep{bramstedt2020}, and it was difficult to know which of the publicly available research could be trusted on any given weekday. This complicated the modelling process, due to direct uncertainty about the disease's pathophysiology, the variable quality of the rapidly collected data and the need to frequently revise the model to incorporate new knowledge (from peer-reviewed and non-reviewed published sources, raw data, or expert opinion). Just digesting and navigating the high volume of new information contributed to the difficulties. All this in turn led to great uncertainty around the scope and purpose of the models and therefore the best modelling decisions, which in itself led to the choice to focus on the causal knowledge base (and its expert models) prior to development of any application models.

\paragraph{Need for novel approaches to the model development process.} 
Given both the uncertainty about, and the potentially changing scope of, the problem domain, the pool of experts who were considered best suited to participate in the elicitation also shifted rapidly. Government-instituted restrictions on movement during the COVID-19 pandemic meant that a larger pool of experts became available than might have otherwise been the case, and the degree of concern about the pandemic motivated many of these experts to support the project. These factors meant the modelling team needed to decide what expertise was most required at any given time. Logistically, the lock-downs forced all elicitation workshops and collaboration (including among the modellers) to be conducted virtually. Fully remote elicitation processes therefore had to be designed with a changing pool of experts; however, this constraint came with unexpected benefits, stimulating new thinking with regard to how elicitation of causal knowledge bases and causal BNs can be more dynamically and efficiently conducted. 

\paragraph{Challenges in defining the end-users} The rapidity with which COVID-19 emerged as a problem meant there was an urgent need for many problems to be solved at once. Research was conducted around the globe that might address those problems, but often without specifying which problems the research would address, or who would be the ultimate end-user for the outputs of that research. In other words, the relevant decisions that would need to be made (and even the decision maker) would change as the understanding of the problem domain evolved. In addition, despite the urgent need for tools to better control and manage COVID-19, the pathway to implement these tools was often unclear. While it is not unusual for research to have no explicit end-user, the urgent need for useful outputs of this research meant the modelling team (researchers and academics, rather than decision makers) had to make concrete forecasts about the best possible uses of any final models and outputs, as well as discover the best possible implementation pathways for those uses. Fortuitously, this led to some very concrete use cases being developed early on, which helped to direct the development of the causal knowledge base.

\subsubsection{Why we pursued a new approach}

Several approaches to developing a BN and even a general knowledge base or ontology for a later BN have been developed or attempted elsewhere~\citep{helsper&vandergaag2002,nadkarni&shenoy2004,korb&nicholson2010a,fenz2012}. The CKE process we describe in later sections draws upon ideas from these approaches, but nonetheless differs in several important ways. Perhaps the most significant is that we aim to provide greater structure and direction to the process of developing the causal knowledge base.

KEBN~(Ch. 10, \citealp{korb&nicholson2010a}) is a structured process, but is broader in its goals than CKE, aiming for an end-to-end process for building application BNs. In the case of COVID-19, the decision to start with the development of a causal knowledge base was motivated by two main considerations. First, we did not yet have access to reliable data to inform the aspects of COVID-19 that we chose to model, but we expected to obtain such data from clinical collaborations in the future. Second, there was little knowledge about COVID-19 itself, and still less around the causal mechanisms, let alone their causal strengths. We needed a record of this limited causal knowledge first, before we could proceed to a useful application model. So the motivations were practical, but after having gone through this process, we now believe it is good practice for many (though not all) situations, which we discuss in Section~\ref{sec:discussion}.

We might also have followed other knowledge engineering approaches that are not causality-specific, such as conceptual models, mind maps and flow charts~\citep{canas+2004,wand&weber2002,moody2005,davies2011}. However, we primarily needed a causal knowledge base in order to understand what interventions might help (and how) prevent the progression of infection to severe disease, as well as what factors might bias or affect testing and diagnosis. With these eventual purposes in mind, we reasoned that developing an ontology comprising numerous non-causal concepts would be of little help. Also, causal BNs are sufficiently capable of representing the non-causal ontology, particularly when sub-models and approaches like patterns and idioms are utilised~\citep{srinivas1994,srinivas1993,koller&pfeffer1997,neil+2000}. Similarly, merely mapping out non-causal associations would lead to models that could not usefully inform later application models for clinical decision support. The reason for this is simple: decision-making critically relies on causal knowledge.\footnote{As \citet{skyrms1982a} notes, this idea is at least as old as Aristotle.}

Variations of approaches that use a knowledge base to support causal BN development have been trialled in the past~\citep{helsper&vandergaag2002,nadkarni&shenoy2004}. CKE has a great deal in common with these approaches, however there are some key differences. First, the models within the CKE are expected to adhere to the requirement of being causal BNs as much as is possible, including the ability to be parameterised directly. This means that some extra modelling effort is required to ensure the models really are causal, are coherent, and work correctly. By contrast, this is not a requirement for causal maps and ontologies for which only the final model is required to be a consistent (and possibly causal) BN.

A key advantage of ensuring the knowledge base models are based upon valid causal BNs is that the knowledge base is thereby checked and validated from the very beginning for consistency with causal reasoning. In addition, causal BNs can be shared directly, and inspected, used and further developed by others. The causal BNs are independent working artefacts. For COVID-19, we took advantage of this to share our work with others, demonstrate and test the concepts and reasoning, and receive feedback wherever we could.

Assuming causal knowledge as a requirement, there were still several alternatives. We could have chosen to develop the causal knowledge base as a set of causal DAGs rather than (partial) causal BNs. While there certainly is value in this approach, in practice, causal DAGs are frequently only useful when developed with concrete data in mind. This is because variables in a causal DAG do not have explicit state spaces and generally cannot be parameterised by experts. Of course, there is no technical impediment --- but if they do have explicit state spaces, or are so parameterised, they are equivalent to causal BNs; at which point, it becomes a matter of terminology what we call them, rather than a difference in representational potential.

There were several alternatives to encoding the causal knowledge, such as conditional probabilistic equations or logic statements~\citep{laskey2008,carvalho+2017} or even plain text. While these can be good options, causal BNs in their graphical form are likely to be more intuitive for experts to understand and navigate, while still ensuring that the causal knowledge is encoded in a form that is internally consistent, both ontologically and logically. Amongst other benefits, this permitted a straightforward qualitative parameterisation and thereby allowed the implications and behaviour of the causal knowledge to be presented back to experts for checking and evaluation.

In addition, causal BNs allow one to integrate information from a variety of sources. While the causal knowledge base for COVID-19 was developed primarily with expert input, we were also able to integrate information from literature and, later, data, and to explore qualitative parameterisations based on the combination of experts and data.

In summary, we chose to follow a CKE process for the following reasons:

\begin{itemize}
\item The ability to combine expert knowledge, literature, existing models and data as it became available
\item A lack of data and pre-existing structured causal knowledge about the problem domain
\item The availability of unstructured causal knowledge held by experts 
\item The suitability of causal BNs for representing causal knowledge 
\item The ability to ensure the knowledge encoded in the knowledge base is causally valid
\item The ability to share causal BNs that can be used to educate others and support and perform causal analyses 'out-of-the-box'
\item The ability for causal BNs to support decision-making, which critically relies on causal inference
\item The ability for causal BNs to eventually provide probabilistic projections given any amount of evidence, helping us to support the later development of application models for applied settings, such as models of COVID-19 prognosis given whatever information is available on signs, symptoms, background factors and comorbidities
\end{itemize}

\section{CKE: A Causal Knowledge Engineering approach} \label{sec:ke}

\begin{figure*}
    \centering
    \includegraphics[width=0.9\textwidth]{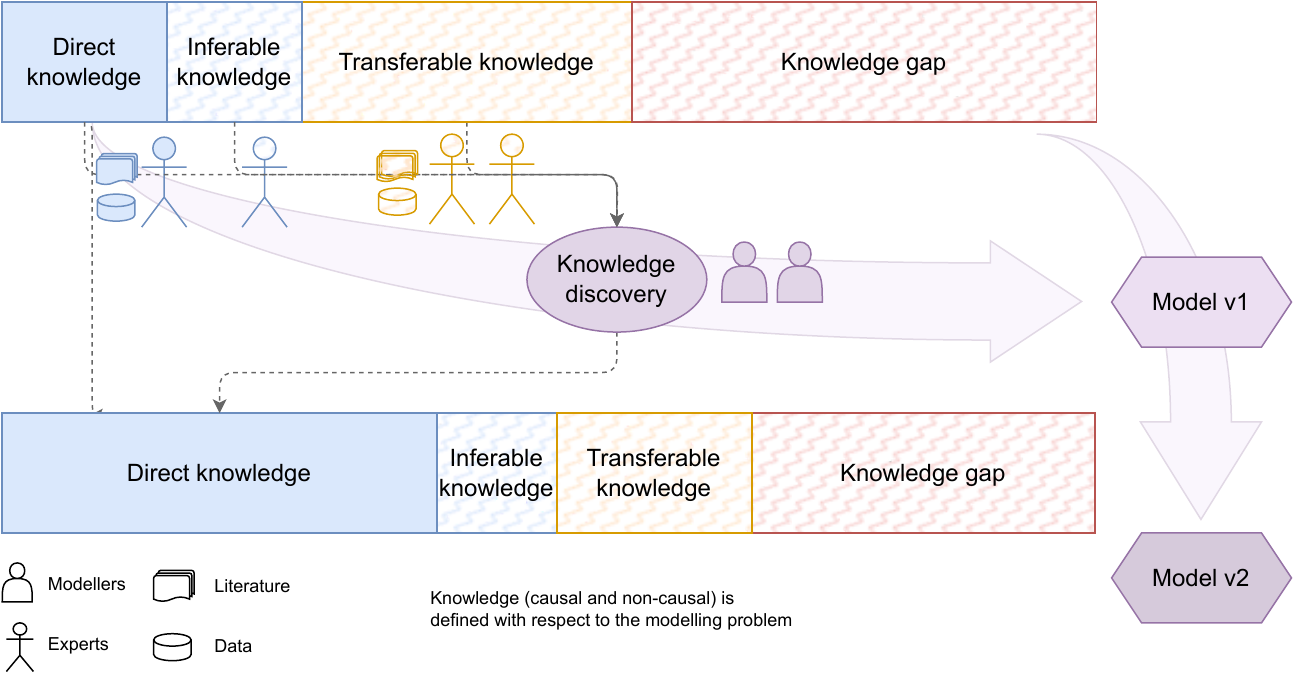}
    \caption{With each iteration, the amount of direct knowledge grows, in turn supporting each new version of the model. Direct knowledge here refers to knowledge (causal and non-causal) directly applicable to the current modelling problem. Transferable knowledge refers to knowledge about other problems or domains that can be applied to the current modelling problem. Inferable knowledge sits in between --- it is knowledge that can be inferred from either the direct or transferable knowledge, that is not currently a part of either. (As a simple example, if it is known that A causes B and separately that B causes C, it may still not be explicitly known or recognised that A causes C.)}
    \label{fig:know_growth}
\end{figure*}

A key motivation for the CKE approach is to grow a base of causal knowledge that can then be used to support application model development (Figure~\ref{fig:know_growth}). Broadly speaking, there are two key groups of people that enable this process: the modellers and the experts.\footnote{We expand on these and further roles in more detail later.} In general, modellers guide the overall process of knowledge discovery and model development, while experts transfer relevant knowledge to modellers, and in turn modellers transfer discoveries from the modelling process back to experts, for the benefit of the experts as well as for validation.

There are multiple ways in which the process of knowledge growth can occur. At any given point in time, there is a base of knowledge available for developing the model along with a knowledge gap in the current modelling problem. We divide the knowledge base into 3 knowledge types, each of which is handled differently. These are direct knowledge, transferable knowledge, and inferable knowledge.

{\em Direct knowledge} is knowledge that is specific and directly applicable to the modelling problem: for example, in our case study, this might be the knowledge that SARS-CoV-2 infection is a necessary cause of the COVID-19 disease that can, in turn, cause severe morbidity and death. Initially, the direct knowledge may be small, as was the case with COVID-19, due to the lack of history and experience with it. Alternatively, the direct knowledge about a problem domain may be considerable, as might be the case with relatively long and well studied diseases such as influenza.

Peripheral to this direct knowledge lies {\em transferable knowledge}. This is knowledge that exists in other domains or about other modelling problems that can be readily transferred or transported to the current modelling problem. For example, in the case of COVID-19, there was a great deal of existing knowledge about human pathophysiology, like the causal mechanisms of pulmonary thromboses, that could be transferred directly to models of COVID-19.

Sitting between both of these is {\em inferable knowledge}. This is knowledge that can be generated by applying inference to direct or transferable knowledge (or both). For example, if it is known (as part of our direct knowledge) that, probabilistically, SARS-CoV-2 infection (I) may cause viraemia (V), and it is also known (as a part of our transferable knowledge) that viraemia (V) may cause direct viral injury (J), we can infer that SARS-CoV-2 infection (I) may cause a direct viral injury (J), and make that a direct part of our causal knowledge base. It's worth emphasising that this knowledge is generated {\em solely} by applying the rule of transitivity --- there may be no observation or experiment at all that includes both I and H. Inferable knowledge may come in many other forms beyond transitivity (e.g., analogy, mutually exclusive causes, impossible scenarios), although transitivity is perhaps the most important.

Direct and transferable knowledge (and, indirectly, inferable knowledge) may come from many different sources, which we group into 4 main types here: existing models, literature, data and experts. Existing causal models that tackle a similar problem in terms of purpose or scope are the most directly relevant sources of causal knowledge. Other less directly relevant models can also be a valuable source of causal knowledge, if care is taken around issues of transportability. Most commonly, knowledge is available as information recorded in the literature. This is typically in an unstructured, plain text format, particularly for causal knowledge. Knowledge may also reside somewhere in recorded or structured data, or associated documentation (which we group with data, rather than literature). There are of course known difficulties with extracting causal knowledge directly from data~\citep{spirtes+1993,korb&nicholson2010a}, but it can nevertheless contain invaluable direct and indirect causal knowledge. Finally, knowledge is also held by experts. This may be the same knowledge that is available in the literature, with experts acting as guide to that knowledge, or can contain unique and unrecorded personal observations or experience in dealing with situations from the problem domain. Experts may have also done a significant amount of causal reasoning about problems from the domain, and therefore possess causal insights that also go unrecorded.\footnote{Knowledge of the domain may also be held by modellers, but in such cases the modellers play the role of experts.}

CKE regulates these processes and sources of information, and in particular, provides a guide for modellers overseeing the process of knowledge discovery and model building.

\subsection{Function and structure of the causal knowledge base} \label{sec:funcStructCkb}

\begin{figure*}
  \includegraphics[width=1\linewidth,page=1]{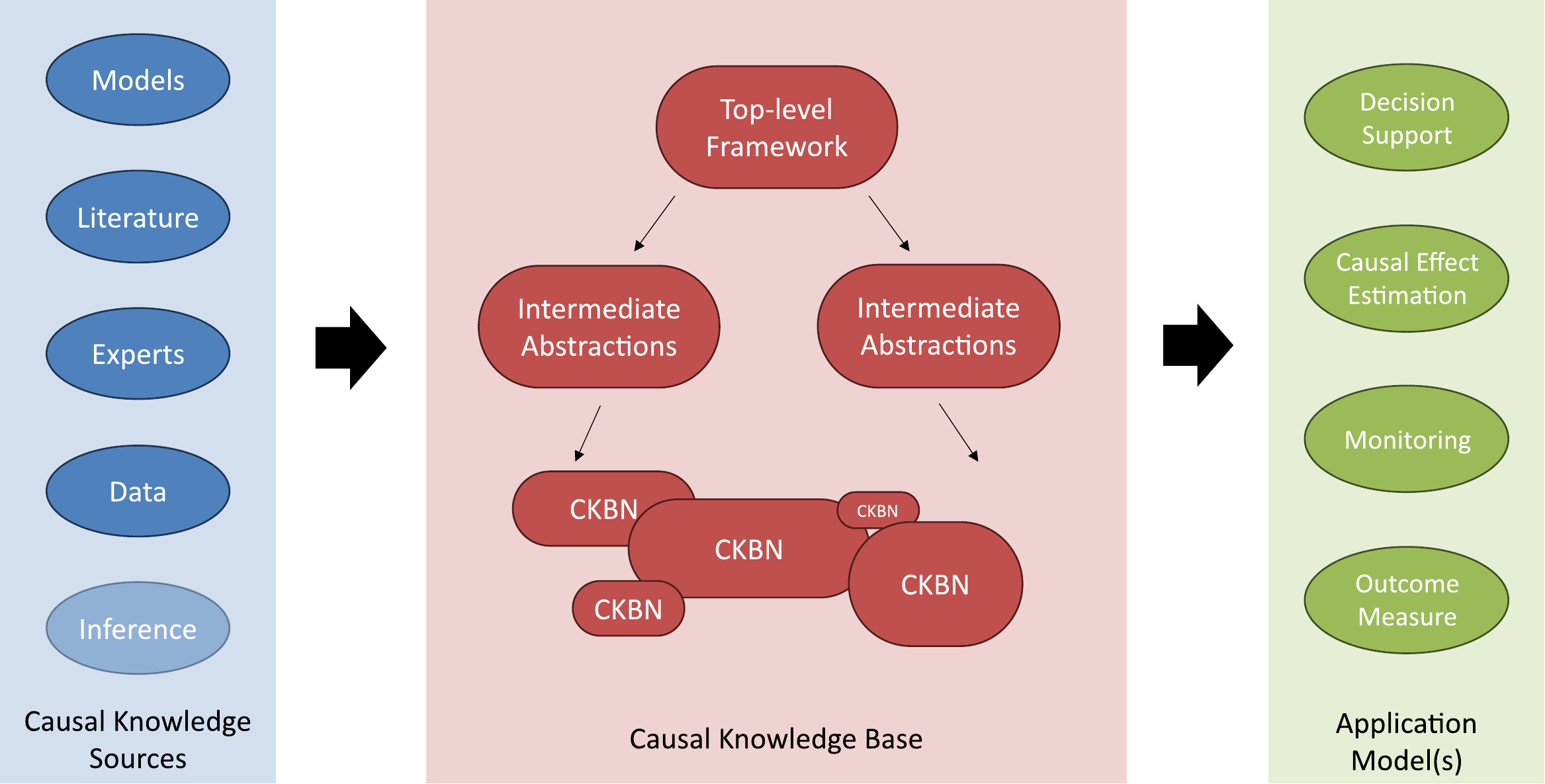}
  \caption{The function (external relationships) and structure (internal form) of the causal knowledge base. We define \textbf{models} to be any existing relevant model (BN or otherwise), \textbf{literature} to be published reports that may or may not be peer-reviewed where the reports may be descriptive summaries, or may include the results of analyses and conclusions, \textbf{data} to be raw recorded values that are not summarised or analysed, and \textbf{experts} to hold beliefs that may, to a varying extent, be informed by the other two or may just be from personal experience and/or extrapolation from related experiences. Inference can be applied in conjunction with any or all knowledge sources, as depicted in Figure~\ref{fig:know_growth}.}
  \label{fig:ontology}
\end{figure*}

The causal knowledge base's function and structure are shown in Figure~\ref{fig:ontology}. The function of the causal knowledge base is to support the development of application models or applications by synthesising and encoding the relevant causal information from disparate causal knowledge sources. Relevant causal information may be drawn from models, literature, experts and data that supports the development of an application. The application may require a purpose-built application causal BN, such as one that might be used for a clinical decision support system, monitoring system or causal effect estimation. The application model may not even be a causal BN; it may be a predictive BN that makes use of causal knowledge in the design, or any other kind of model, such as a systems dynamic model or one or more linear equations. Alternatively, the application may be informed by the causal knowledge base without any purpose-built application BN. For example, the causal knowledge base may be used to design and distribute an instrument for outcome measurement in a way that minimises bias in data collection. The final application's form may be even more distantly related to the causal knowledge base, such as a data analytics and visualisation dashboard, or a generic software application where the causal knowledge base captures relevant knowledge of how users interact with the software.

The structure of the causal knowledge base can in principle take on any form suitable for the problem domain and for supporting the final application. For the CKE process, we recommend that the causal knowledge base be structured as a hierarchical set of models. This resembles an object-oriented BN approach~\citep{koller&pfeffer1997}, however provides for more flexibility, allowing overlaps in individual CKBNs, alternative definitions across different models and multiple resolutions. The model at the top level provides an overall framework for the knowledge base, and is called the top level framework. An example of the top level framework developed during the COVID-19 CKE is shown in Figure~\ref{fig:toplevel} of Appendix \ref{sec:examples}. Each lower layer in the hierarchy elaborates on some piece of the layer above. This does not need to be done comprehensively. For example, in the COVID-19 top level framework, a node for epidemiology appears. However, at no point was our team intending to produce (transmission) models of COVID-19 epidemiology.

Below the top level framework lie the CKBNs. Table~\ref{tab:ckbn_list} provides a list of the specific CKBNs that were developed for the COVID-19 causal knowledge base. These may appear in any number of layers as would be convenient for the modelling problem, with shallow hierarchies preferred and growing to deep hierarchies only as a way to organise a larger causal knowledge base. The COVID-19 causal knowledge base effectively contained just a single layer below the top level framework, although there were several hierarchical relationships between these CKBNs that could have been made explicit. For example, the Respiratory and Complications BNs (\citealp{mascaro+2023}) could have been re-organised into separate layers, with the Complications BN being generalised to COVID-19 pathophysiology at the organ level, and the Respiratory BN being one of the organ systems.

Every element (e.g., top level framework, CKBN) of the causal knowledge base should have its own general documentation that describes its purpose and how it was created. Every element should also have a dictionary, which is essentially documentation for each element inside the framework or causal BN. For COVID-19, we provided dictionaries for each individual CKBN but not for the top level framework. At the time, the framework was only used for illustration and to facilitate communication, such as during workshops, but we would now consider it a formal part of the knowledge base. 

\subsection{Roles}

We define a number of key roles involved in the development of a causal knowledge base. One person can fulfil multiple roles, and multiple people can share the same role. Roles include the following:

\begin{itemize}
    \item {\bf Causal knowledge base lead:} Has stewardship of the causal knowledge base, ensuring that it meets the purposes of any known or potential application models
    \item {\bf Problem owner}: Responsible for defining the problem. The causal knowledge base can support understanding multiple problems, each with its own problem owner. The problem owner helps with identifying applications, scoping, assisting modellers as needed with planning, management and resourcing, and ensuring that existing work is made available
    \item {\bf Application lead:} Responsible for delivering an application designed to solve a problem that will be based on application models derived from the causal knowledge base. Works with the causal knowledge base lead to ensure the knowledge base fulfils the requirements of the application
    \item {\bf Resource provider:} Provides resources (such as funding, software, equipment and access to experts) for the development of the causal knowledge base
    \item {\bf Stakeholder:} Anyone who may be affected by the eventual use of the product, including end-users and potentially the general public. Stakeholders may or may not be directly involved in the model development, depending on the purpose
    \item {\bf Modeller:} Supervises all or part of the CKE process that involves knowledge discovery and model development that form part of the causal knowledge base. There are several types of modeller, including:\begin{itemize}
        \item {\bf Modelling lead:} Responsible for final decisions on the model development, and in particular what is to be included in the causal knowledge base
        \item {\bf Model developer:} Contributes to the formalisation of causal knowledge into causal BNs
        \item {\bf Documentation developer:} Provides or reviews documentation for the causal knowledge base
        \item {\bf Elicitor:} Elicits knowledge from experts. This may or may not be someone with formal causal BN expertise, but is still a part of the modelling team
        \item {\bf Facilitator:} An elicitor who facilitates knowledge elicitation from expert {\em groups}, typically in synchronous settings
        \item {\bf Expert co-ordinator:} Assists with recruiting and communicating with relevant experts for elicitation workshops and 1-on-1s. May not have formal causal BN expertise
        \item {\bf Reviewer:} Assesses models and the broader causal knowledge base in terms of technical correctness and internal consistency. (Note that this differs to an {\em expert} reviewer's focus.)
    \end{itemize}
    \item {\bf Expert:} Has knowledge, or knows how to obtain knowledge, about the modelling problem. As with modellers, there are multiple types of expert: \begin{itemize}
        \item {\bf Advisory expert:} An expert with high availability, who the modelling team can consult with regularly during the CKE process. The advisory expert can advise on expert selection and help with recruitment, assists with initial validations and aids the modelling lead in making final model decisions
        \item {\bf Reviewer:} Assesses (typically individual) models for the degree to which they capture the intended target systems that are being modelled. (Note that this differs to a {\em model} reviewer's focus.)
        \item {\bf Literature expert:} An expert who can provide knowledge of which literature to consult, or who can directly extract the relevant causal knowledge from literature
        \item {\bf Data expert:} An expert who is familiar with not just available datasets, but also the structure and definitions of data, and the value and relevance to the modelling problem
        \item {\bf Practice expert:} An expert who can provide knowledge from their own direct experience and practice in the domain
    \end{itemize}
\end{itemize}


\subsection{Workflow}

\begin{figure*}
  \centering
  \hspace{2cm}\includegraphics[width=1\textwidth]{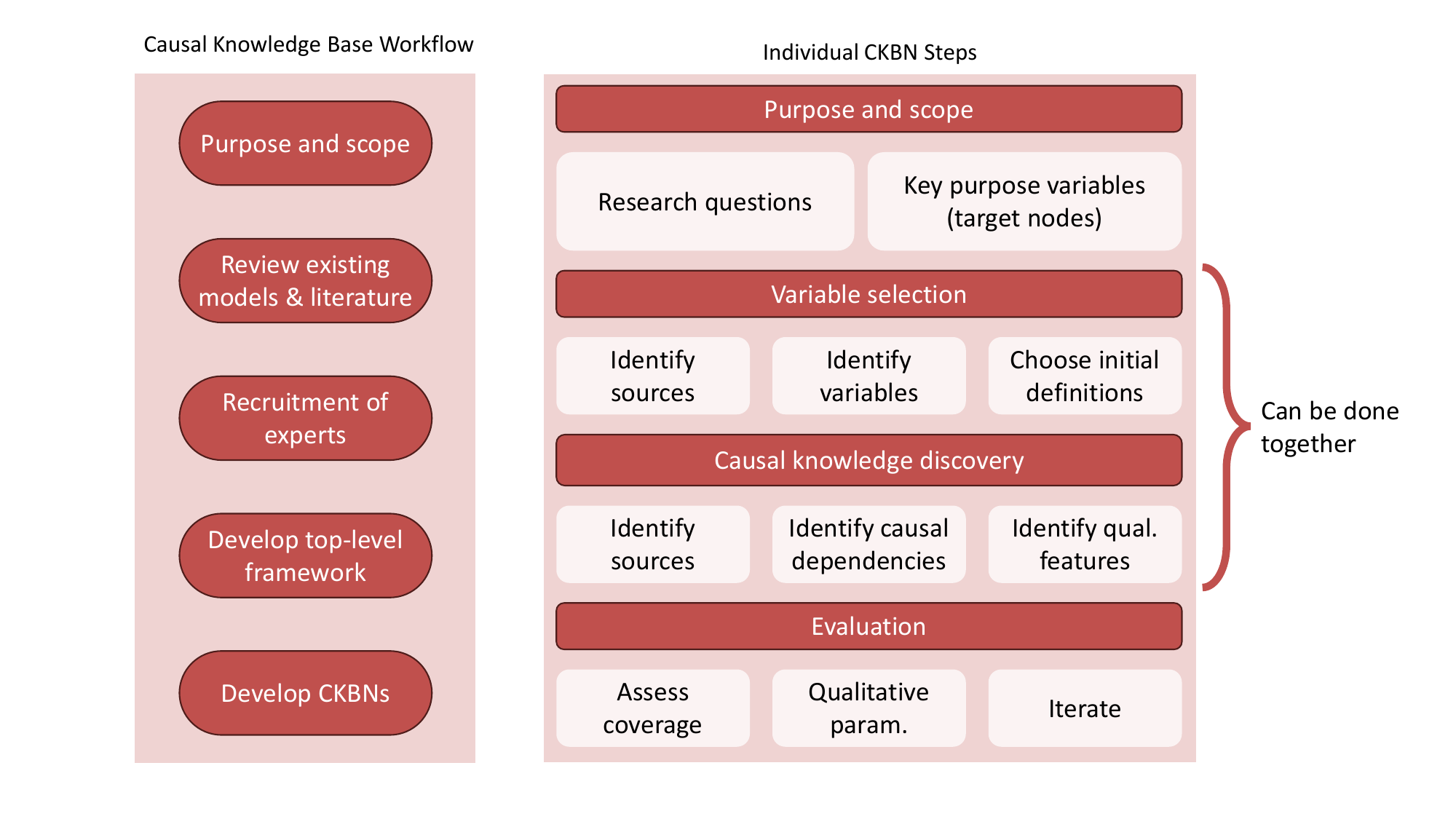}
  \caption{Workflows for the causal knowledge base (left) and and individual CKBNs (right). The workflow for the causal knowledge base proceeds through 5 major steps, starting with the purpose and scope, proceeding through reviews, expert recruitment and then development of the causal knowledge base itself (the top-leve framework and all the constituent CKBNs). The development of an individual CKBN is initially similar, focusing on purpose and scope, and then proceeds through the familiar stages of causal BN development; here we provide additional recommendations on what these stages might contain.}
  \label{fig:elicitation_process}
\end{figure*}

At the highest level of abstraction, CKE can be mapped out as a workflow (Figure~\ref{fig:elicitation_process}). Due to the extremely open, creative and subjective nature of creating CKBNs that will depend on the problem domain and the use cases, it is not possible to prescribe a single clean workflow that describes each step in exact detail. For each step of the workflow, a number of possible alternative and complementary approaches may be useful depending on the problem. While we make no attempt to map these out here, it is nonetheless possible to provide a general outline along with some key options, particularly focusing on those we applied during the COVID-19 work. This workflow provides guidance, particularly to novice causal BN modellers, and a way forward for producing causal BNs that will be fit-for-purpose. Indeed, it is with the intended purpose that the CKE workflow begins.

\subsubsection{Purpose and scope} \label{sec:ckbpurpose}

Defining the purpose of any model is critical~\citep{chen&pollino2012}, and while a causal knowledge base does not require an immediate and sharply defined purpose like an application model, it is still necessary to have one that is clearly understood by all. This may just be to support the eventual purpose of any future application model, or may be broader. The purpose can often be a simple explicit statement that is referred to regularly and used to guide the CKE process. For example, in the COVID-19 work, the very first statement of the modelling purpose was of the following form:

\begin{quote}
COVID-19 decision support for:
\begin{itemize}
    \item In-hospital prognostication for COVID-19 patients
    \item Home-based monitoring for confirmed cases (managed by GP)
    \item Diagnosis (by GP or via other telehealth service)
\end{itemize}
\end{quote}

This statement was formulated by the modelling lead early in the modelling process, and based on discussion between the modellers, advisory expert and causal knowledge base owner, and accepted by the resource provider. It was reviewed at the beginning of every group workshop session and for many of the 1-on-1 sessions. It evolved over the course of subsequent months, as we discovered new opportunities, and found other opportunities to be infeasible. It was also formulated at an early stage when we expected the application models to be direct, refined versions of the elicited models. However, much of the basic purpose remained the same throughout the longer project, and this purpose guided the way in which we developed the causal knowledge base. Of course, the purpose here is for the final application and associated application models. Had we had a formal causal knowledge base in mind, the purpose would have been written to more clearly state the need for it to support all the various types and varieties of application models that we could foresee; however, for the purposes of later sections, we take this to describe the purpose of the COVID-19 causal knowledge base, not the subsequent applications.

Note that the COVID-19 CKE purpose mentions a set of planned uses, that can be considered to be coarsely defined use cases. While this was unintentional, we have come to believe that explicitly elaborating the use cases of the expected applications is of key importance, which can in turn inform the purpose, as well as guide and influence, the development of the causal knowledge base. While we did not develop use cases at the time of developing the COVID-19 causal knowledge base, we did so later as we worked on a user interface for decision support. Examples of some use cases for the prognosis application are as follows:

\begin{enumerate}
    \item Provide clinicians with a high-level assessment of a patient's risk of severe outcomes, so that they can make decisions about alternative interventions or management strategies
    \item Provide clinicians with detailed information on the disease status of individual patients, so that they can make specific treatment and intervention choices
    \item Educate students and junior clinicians on the typical features, risk factors and indicators of COVID-19 pathophysiology, and how they affect outcomes
    \item Provide ward and hospital managers and government staff with high level estimates of patient disease statuses and potential treatment requirements, so that they can better allocate resources
\end{enumerate}

Alongside the purpose is the scope, which is often constrained by the resources available for the project. Typically, the resource provider and other owners would be a party to the discussion of scope. It is inappropriate, for example, to build a comprehensive COVID-19 causal knowledge base if there is likely only funding and time for one tightly focused COVID-19 application. In our case, COVID-19 presented special circumstances, and the resource provider gave leeway for the scope to be decided by the modelling team, based on what the team believed would best suit the purpose within the available funding and time. Use cases, if available, can of course guide decisions around scope to an even greater extent.

\begin{figure*}
    \centering
  \includegraphics[width=0.7\textwidth]{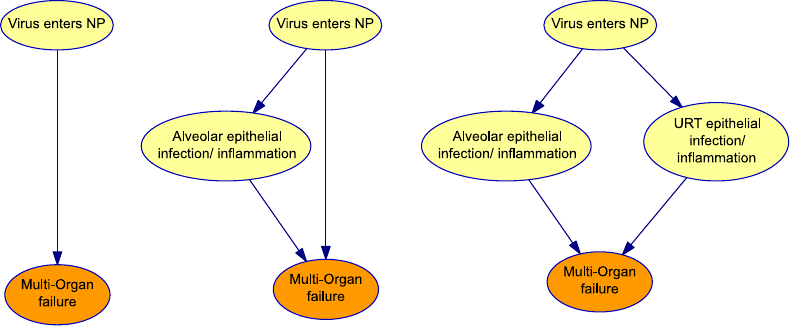}
  \caption{Multiple causal BNs for modelling the relationship of Virus enters the NP (Nasopharynx) to Multi-Organ Failure, in addition to the other (major) causal BN (the full Respiratory BN) from~\citet{mascaro+2023}.}
  \label{fig:multi_multi_organ}
\end{figure*}

Note that the open, creative and subjective nature of creating a causal knowledge base refers to its specific content and form --- not to the question of whether it fulfils a particular purpose or set of use cases. This poses a problem for decisions around scope, as there are {\em many} different ways the knowledge base can be formed that can fulfil a particular purpose (i.e., the form is underdetermined by the purpose), and typically the question of what particular content and form is used depends on external factors such as available time, appetite for complexity, clarity, generality, and intended future uses and reuses. The issue of an underdetermined form is common to other mapping (e.g., concept mapping) techniques (see \citealp{davies2011} for example). As a simple example, Figure~\ref{fig:multi_multi_organ} shows multiple ways that the causal relationship between Virus enters NP (Nasopharynx) and Multi-Organ Failure may be represented in the causal knowledge base. Assuming the general correctness of each, there is no {\em independent} reason to prefer any one over any of the others. Simplicity is typically argued for, but unless we have a very concrete sense of current and future purposes, we cannot know what level of simplicity is appropriate. If, for example, we only expect interventions to be available to prevent a virus from entering the NP, then perhaps just the first model is suitable; however, if we additionally expect it to be possible to intervene to prevent alveolar epithelial infections, then perhaps the second model may be more appropriate. Further, suppose we are {\em currently} unaware of any way to intervene to prevent upper respiratory tract epithelial infections; such interventions may nonetheless be possible in future. The issue here is similar to that for deciding on features during software development, but felt more keenly due to the potentially high value and deceptively low cost of adding each piece of additional causal knowledge.

These decisions affect not only what we include or exclude from the knowledge base, but also how we 'carve up' nature. We might choose to merge infection and inflammation or keep them separate. Or we may decide that the upper respiratory tract is less useful to model than the first three quarters of the respiratory tract. The art of carving nature may rely less on capturing the world precisely, and more on using whichever concepts experts are familiar with, in which the literature communicates, and in which data is commonly recorded. 

\subsubsection{Review of existing models and literature}

A key task when developing a causal knowledge base is consulting existing work, including existing literature and models, for relevant causal knowledge. This can be undertaken by the modellers independently, or guided by experts. The advisory expert is particularly important in either providing good guidance about existing work, identifying other experts or perhaps co-ordinating research assistants or students who can assist. Literature reviews can be conducted prior to engagement with the broader expert group, or as causal BNs are formed. Ideally, all key causal knowledge that appears in the knowledge base (that is not inferred from other knowledge) should be traceable to sources in the literature or other models. 

Existing work need not come from far afield. For example, prior to the onset of the pandemic, our team had developed causal application models for infectious disease, with the most established one being a model for paediatric pneumonia. As little was known about COVID-19 initially, the pneumonia BN formed the starting knowledge base for subsequent knowledge discovery. A strong argument for adapting the pneumonia model was that it was in the process of being extended into a generalisable framework for infection. As our team closely followed the literature and WHO releases leading to the general declaration of the pandemic, it was clear that significant differences would need to be accommodated into the specific COVID-19 causal BNs.

\subsubsection{Recruitment of experts}

Unfortunately, it is rare that literature and existing models are sufficient for building the causal knowledge base. Even if all the necessary knowledge could be found in existing work, it can be difficult for modellers (even with some familiarity of the area) to easily navigate and extract that knowledge. Hence, experts play a crucial role in surfacing that knowledge~\citep{burgman2016a,ohagan2019}. Experts also fill in the gaps by providing evidence from their own direct experience and using their experience to make inferences that are important to fulfilling the purpose of the knowledge base.

Just as with the modellers themselves, there are also issues with bias amongst the experts~\citep{burgman2016a}. Some of this bias can be mitigated at the recruitment stage by ensuring there are multiple experts and that they constitute a diverse group, not just in terms of their field of expertise, but also background and culture. There are many good discussions of the importance of diversity in the expert elicitation processes~\citep{burgman2016a,hanea+2016,ohagan2019}. The number of experts needed is difficult to estimate --- as described later, we have found a good minimum to be 3 for individual CKBNs, which is similar to minimum estimates made by others~\citealp{crispen&hoffman2016}. So a rule of thumb for the causal knowledge base overall might be as many experts with the right fields of expertise, and ensuring that each CKBN has at least 3 experts that can be consulted.

Identifying the required fields of expertise can be guided by the top level framework (discussed in the next section) and the advisory expert. The actual methods for recruitment will be very problem dependent, but convenience sampling and snowball recruitment (through the full project team, and particularly product owners and advisory experts) can work well if it can cover the required fields of expertise and meet the diversity requirement. Of course, today there is a wide array of alternative recruitment methods~\citep{wasfi+2021}. Identifying communication channels (such as online sources, networking events and conferences) where experts are likely to be engaged and interested in the problem domain is crucial.

The COVID-19 pandemic afforded novel (and unlikely to be repeated) opportunities for our modelling efforts, as it strongly motivated experts to support the project as well as provided them with more availability than they would otherwise normally have. Early research suggested that COVID-19 was manifesting in the respiratory system. As a priority, we sought clinical experts in respiratory and infectious disease medicine. Cardiology and renal expertise were also of interest as it had been reported in the literature that these organs were affected in severe cases of COVID-19.

Experts were primarily convenience sampled by leveraging the project team members' existing networks. Snowball recruitment was also employed, with experts encouraged to recommend appropriate individuals from within their own networks. Email invitations were sent out to known contacts, some of whom were members of a national email-based discussion forum for infectious diseases physicians who subsequently provided access to the distribution list. Experts were considered eligible if they expressed interest in participating in elicitation activities.

The worldwide lockdowns made remote participation mandatory. While this presented new challenges, it also provided many benefits. Experts came from a wide geographic area, and could participate with much greater convenience and lower cost --- allowing more experts to participate, and on a more regular basis. Modellers were therefore able to have flexible interactions with experts, with elicitation activities conducted in shorter and more regular sessions of 1-4 hours, rather than concentrated 1-2 day workshops.

\subsubsection{Top level framework}

The causal knowledge base is developed in an iterative fashion. The first stage in its development involves mapping out a top level framework that covers the problem domain at a very high level of abstraction. This can be done in parallel with the initial literature review and recruitment of experts as it can both guide, and be guided by, those processes. Ideally, the top level framework will map out in abstract form the key inputs that should be accounted for by the causal knowledge base, along with the key outputs and decisions or actions that are taken in response to the outputs. In addition, the top level framework should have sufficient scope to cover everything in the causal knowledge base. There may be cases in which this is not possible, and hence multiple frameworks are required instead --- again, these should jointly cover the scope of the causal knowledge base. The top level framework should be causal, although the numerous interactions and feedback loops between the components of the framework may sometimes render the causal interpretation of limited value.

Once developed, the top level framework can be used to identify whether the intended scope is appropriately covered by other CKBNs in the causal knowledge base. It may also be used to help identify areas that are explicitly outside of the scope.

An example of the top level framework developed during the COVID-19 CKE is shown in Appendix~\ref{sec:toplevelframework}. The framework remained relatively constant throughout the CKE process and did not explicitly include any feedback loops, though these are certainly possible if treatment effects are incorporated into the process. Being sufficiently general, the framework was also later adapted for guiding the design of the prognosis application model.

\subsubsection{CKBNs} \label{sec:ckbn}

Below the top level framework lie the individual CKBNs. There may in fact be multiple levels in the hierarchy, starting with the top level framework, moving down to high level abstract models, and then finally to detailed causal knowledge models. In the case, of the COVID-19 models, we had two layers --- the top level framework in the top layer and CKBNs at a single level of abstraction in the bottom layer --- and we assume here a CKE with these two layers.

The ways in which a CKBN in the bottom layer can be developed are numerous. We outline a broad array of the individual techniques and options in Appendix~\ref{sec:building}, particularly considering those used during the COVID-19 CKE, with descriptions of how they were (or were not) used in the case study. Due to their variety, we do not discuss individual techniques in this section, and focus instead on the high level workflow for developing a CKBN, which will almost always have the same general structure.

\paragraph{Purpose and scope} The first step in the development of a CKBN (as with the causal knowledge base overall) is defining its purpose and scope. In this case, however, the purpose and scope is specific to the role it will play in the larger causal knowledge base, which may be determined directly from its location in the top level framework. For example, in the COVID-19 CKE process, one of the CKBNs developed focussed on the respiratory system. The purpose of this model was to identify how infection with SARS-CoV-2 could lead to problems in the respiratory system, with the goal of identifying features of the system that could be affected by background risks,  intervened upon through treatment or therapy, and which could be observed or measured.\footnote{Note, the goal was {\em not} to identify what those background risks were, what the treatments and therapies might be or what the specific measurements might be.} The scope, along with key inputs and outputs for this CKBN, was largely settled by the top level framework (i.e., the COVID core mechanism submodel).

\paragraph{Key purpose variables} The most important variables that are required for fulfilling the model purpose should be identified next. This might just be one or two variables, or it might be an entire class of variables. As an example of the former, in the COVID-19 respiratory model, we considered the key variables to be SARS-CoV-2 virus entering the nasopharynx, and multi-organ failure (or otherwise death). The omission of either of these variables would have made it difficult to stay within scope. For instance, multi-organ failure might be caused by factors having nothing to do with SARS-CoV-2; and SARS-CoV-2 might have any number of effects that are of no interest for clinical prognosis. By contrast, variables that do not mediate the link between virus entering via the nasopharynx and multi-organ failure can be easily recognised as needing to be omitted, providing a very natural way to scope the model in terms of subject matter. This does not provide any help with scoping in terms of breadth and depth --- i.e., how many mediating pathways and how many steps in each pathway should be modelled, which is considered next.

Instead of, or in addition to, specific key variables, one may settle on one or more key classes of variables. For example, it may be that the goal of the CKBN is to identify all background factors for some disease, or instead all symptoms. Hence, the CKBN's purpose can only be achieved to the degree that a comprehensive set of such factors (or symptoms) are identified. 

\paragraph{Variable selection} With the purpose and subject matter scope settled, an initial selection of variables can be performed. This can be done in many ways, and the specific approach to variable selection needs to be designed. This involves identifying: the sources that will be utilised (e.g., experts, literature, data), the way variable extraction will be performed, the types of variable that are of interest, and possibly the approximate number of variables that are estimated to be sufficient. In addition, variable definitions need to be selected or specified, particularly where there is a lack of clarity around those definitions.

There may be no, partial or complete overlap with the next step in which causal knowledge discovery is performed. For example, there may be no overlap if the source for variable selection is a specific dataset, and there is only the intention of drawing the causal relationships between the variables in the dataset, not of introducing latent or summary variables. At the other extreme, it may be decided that no variables (aside from the key purpose variables) will be selected ahead of time, and that they will all be derived via the causal knowledge discovery process. The latter is most similar to the process we followed during the COVID-19 CKE for the Respiratory BN.\footnote{While we did have an initial template for the Respiratory BN, only a few elements were kept from that template given the experts' belief in the novel pathophysiology of COVID-19.}

\paragraph{Causal knowledge discovery} Identifying causal relationships is the final step. This will ideally involve both a) identifying variables that are causally related and b) identifying {\em how} they are qualitatively causally related (e.g., does the presence of the cause {\em increase or decrease} the probability of the effect). If good data is available, it may be possible to use causal discovery (machine learning) to identify causal structures from data~\citep{spirtes+1993,wallace&korb1999,odonnell+2006}. However, in most cases, experts will need to either drive or be heavily involved in the process. Again, there are an array of specific techniques that can be used and hence the approach needs to be designed. For example, it may be decided that experts will be used as a source of knowledge, that a set of group workshops followed by 1-on-1s is the preferred format to elicit that knowledge, and that the reverse bow-tie method will be used during the group workshops and validation and elaboration used during the 1-on-1s. In the COVID-19 CKE process, this was the main approach for the development of the Respiratory BN.


\paragraph{Qualitative parameterisation} An optional, but recommended, component of the workflow for developing a CKBN is to provide a {\em qualitative parameterisation}~\cite{mascaro+2024a}. A qualitative parameterisation is one that captures the qualitative nature of the causal relationships (e.g., when the probability of virus entering NP increases, so too should the risk of upper respiratory infection), without making any claims to quantitative validity or precision. This can be done to various states of completeness, for example: state spaces assigned to variables; discretisations; modeller provides parameterisation based on logical relationships; modeller provides parameterisation based on estimated prevalences and conditional probabilities; individual experts provide independent estimates; or experts provide more precise conditional estimates. In principle, it is also possible to use data as part of the qualitative parameterisation process, but in practice it is difficult to separate out the qualitative component from the quantitative component when doing so.

Qualitative parameterisation of any causal BN (whether as part of a knowledge base or otherwise) is best thought of as a way to {\em illustrate} the intended behaviour of a model, rather than a rigorous attempt at parameterisation. It may be helpful for experts to sense check the behaviour, but in general there is usually little purpose in (for example) performing rigorous validation against data.

\paragraph{Evaluation of coverage and iteration} Finally, evaluations should be performed on the developed CKBN to see if it achieves its purposes, and contains an appropriate level of detail. If this is not the case, then further iterations may need to be done.

\subsubsection{Techniques for developing CKBNs} \label{sec:techniques}

It is difficult to provide a clear and rigorous workflow for the development of a CKBN. As with the causal knowledge base itself (Section~\ref{sec:ckbpurpose}), developing an individual CKBN is a similarly open, creative and subjective --- but again, whether the purpose has been achieved is not subjective --- and what works well for one problem, modelling team and expert group may not work so well for another. Rather than give a prescriptive workflow that is of little practical use, or otherwise give up on the goal of providing any guidance entirely, we can provide useful guidance by collecting together the techniques that we found helpful during the COVID-19 modelling effort (and in some cases, other work), along with some recommendations on which techniques work well or not so well under what circumstances.

We catalogue the techniques that we have used in the COVID-19 case study in Appendix~\ref{sec:building}, describing where and how we have used them. We divide these techniques into 3 different categories that approximately correspond with the main phases of CKBN development: utilising causal knowledge sources, developing the CKBN structure, and checking and revising the CKBN.

\section{Eliciting a CKBN from experts} \label{sec:elicit}

Typically, expert elicitation is the most difficult and complex part of the CKE process. There are an array of expert elicitation techniques that can be used (as is discussed in Section~\ref{sec:experts}), and many different ways in which these techniques can be combined. A good approach will depend heavily on the context of the modelling problem. However, we believe it is worth outlining an approach for a hypothetical single CKBN that should work well in a broad range of settings, based on the path we chose in our case study.

\begin{figure*}
    \centering
    \includegraphics[width=1\textwidth]{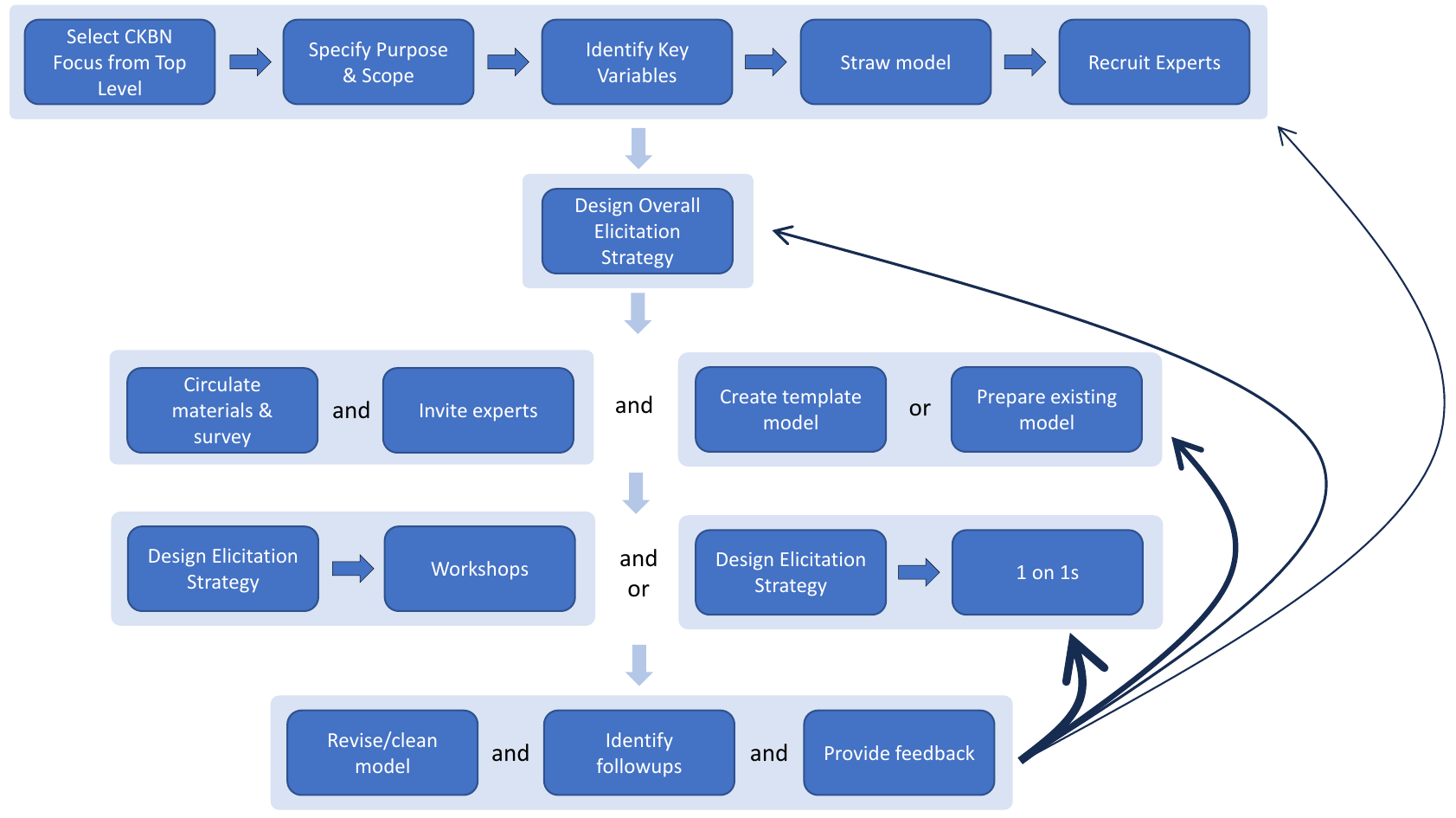}
    \caption{A typical CKBN elicitation workflow. (This workflow focuses specifically on expert elicitation of a CKBN; compare to Figure~\ref{fig:elicitation_process}, which applies to the development of a CKBN using any causal knowledge source(s).) There are 5 main stages, which consist of general CKBN elicitation preparation, design of the elicitation strategy, development of materials for workshops, conducting workshops and revisions and followups.}
    \label{fig:elicit_workflow}
\end{figure*}

This CKBN elicitation approach is given in Figure~\ref{fig:elicit_workflow}. It focuses on group workshops for different key components of the top level framework, coupled with supporting 1-on-1 discussions with experts, surveys and email exchanges. This was very close to the approach adopted in the case study. That approach was chosen for several reasons: 1) early in the pandemic, good information and data around COVID-19 was scarce, but there was a great deal of background knowledge that was likely to be transferable and we needed the help of experts to identify that knowledge; 2) there was no one person who had authoritative knowledge of the required aspects of COVID-19; and 3) our group had access to a large number of experts willing to assist and who each understood different aspects of the likely pathophysiology of COVID-19, and we expected there to be beneficial synergies by allowing the experts to interact. While the considerations for other projects may differ, we believe there are many projects that share at least some of these considerations.

The main workflow adheres to the basic flow that we recommend for developing a CKBN in Section~\ref{sec:ckbn}, involving selection of a CKBN from the top level framework, defining purpose and scope, identifying variables, model structure development, evaluation and iteration. But where we allowed for an array of different techniques in the general CKBN workflow, here we advocate for a particular set of techniques, which we utilised and found beneficial for the COVID-19 case study.

The first step of the CKBN process is to identify the next aspect of the top level framework for which a CKBN will be developed. This should be a decision made by the modelling lead and advisory expert, with support from any other members of the modelling team, and should consider the current state of the causal knowledge base and its intended development path, as well as the way in which the knowledge base supports the purposes of the application models.

As with any CKBN, the next steps are to clarify the purpose and scope and then select key purpose variables. Again, these are tasks that can be done primarily by the modelling lead and advisory expert, requesting support where required. The next steps are also similar to ordinary CKBN development, however we recommend a specific set of techniques that we applied to the various models in our case study. The workflow will differ in focus depending on whether the CKBN is mostly new or existing.

For a new CKBN, the modelling team (with support from the advisory expert) first develops a straw model so as to understand the needs of the CKBN. For instance, the straw model may come together easily and exhibit a clear separation of concepts, or it may suggest work is required to elaborate the overlap across concepts, highlight contradicting ideas or surface other complexities. It may also be clear that some variables in the straw model are more certain than others; such variables can be used as part of starting templates that can be discussed 1-on-1 with experts or used directly in a group workshop. In some cases, if the modelling team decides they have high confidence in the straw model, it may itself serve as the starting model for further elicitation.

The next step is to use the knowledge gained thus far to recruit experts for this particular CKBN and decide how they will best contribute. If there is already a pool of experts who have agreed to contribute to the general project, this should serve as the first candidate list, selecting those who may have relevant expertise for the specific CKBN. If this is considered a sufficient number, no more work need be done; if not, the advisory expert, and current candidates on the list can also be approached to identify others who may be willing to participate. Importantly, experts will have varying availabilities, or otherwise they may leave the project, be difficult to coordinate with, or represent too little diversity. For all of these reasons, the goal should be to reach out to a slightly larger number of experts than required. For example, if the modelling team believe that 4-6 experts would likely provide good (perhaps close to saturated) coverage and diversity (and there is some evidence that this might often be a good number~\citealp{crispen&hoffman2016}), the goal should be to contact at least 6-8 experts for their involvement in the process.

If during development of the straw model and recruitment, there remain some open questions or uncertainties that the advisory expert cannot handle, one or two preparatory 1-on-1s can be held with an additional selected expert or two to see if these can be resolved before taking the elicitation to a group setting.

Once the modelling lead decides everything is ready, the next step in the workflow proceeds to designing the overall elicitation strategy. The strategy needs to be flexible to accommodate different expert availabilities, and involves deciding upon suitable elicitation formats. This includes deciding on: 
\begin{enumerate}
    \item whether to hold group workshops, 1-on-1s, or a mix of both; 
    \item how much will be done online compared to in person (see below for considerations);
    \item whether to hold one or two large workshops, or a sequence of smaller elicitation activities; 
    \item whether to use an open format with large expert contributions, or if experts should be reserved for validation of the modelling team's work; and
    \item whether to develop a model driven by just 1 or 2 experts (potentially the advisory expert), or one that is a product of many minds.
\end{enumerate}
A rough schedule for the elicitation can then be designed. Group workshops (and other difficult to coordinate activities) should be entered into the schedule first, taking into account the work that will be needed in preparation for them and the work needed between workshops. 1-on-1s can generally be scheduled as needed, but ensuring that there is enough time for them in amongst other elicitation and modelling activities. 

It is worth highlighting some of the differences between online and in-person activities. Compared to in-person activities, we found that online workshops worked extremely well during the case study, with many benefits. These included:
\begin{enumerate}
    \item access to a much greater pool of experts;
    \item much lower expenses;
    \item less time required from all parties;
    \item a more consistent approach to expert input;
    \item experts could remain in their work or home environments;
    \item experts had access to their own equipment and had access to supporting materials;
    \item the facilitator had significant control of attending experts' attention, more than is possible in person; and
    \item individual and group follow-ups could be organised more easily.
\end{enumerate}
The primary advantage of in-person workshops is that it allows experts to engage with each other in a free flowing exchange, either within the workshop, or during the breaks and gaps of an in-person get-together. This can allow for greater creativity and foster collaborations that might not otherwise occur. Whether these are important will depend on the problem. For the case study, we found the benefits to online elicitation far outweighed those of in-person elicitation, to the extent that we still primarily conduct elicitation sessions online today.

Once the overall elicitation strategy has been designed, several activities can proceed in parallel. Invites for participation can be sent to experts. The modelling team can also circulate preparatory materials with the invitations, or separately as a follow-up, as is common with any kind of elicitation (see for example \citealp{europeanfoodsafetyauthority2014}). These materials should include: 1) the goals of the elicitation; 2) a quick overview of how causal and Bayesian network modelling works; 3) an overview of the current state of the model or the modelling team's understanding; and 4) a request for literature and other materials (and possibly other experts) that may be of value.

At the same time, the modelling team may prepare a survey and circulate it specifically to the invited experts, or potentially to a broader expert group. Such a survey can be (and often needs to be) brief, and its main purpose is to identify issues that may arise and to set the mind-frame for the elicitation process that follows. The survey should focus on the key questions that would put workshops and 1-on-1s at risk of a poor outcome if the modelling team has made incorrect assumptions. The straw model should be revised based on the survey responses, or otherwise annotated to indicate areas that may benefit from more expert attention.

Just as there is a strategy for the overall elicitation, there should also be a strategy for each individual workshop or 1-on-1. Both the straw model and survey responses can be used to influence the design of these elicitation strategies. Generally, the straw model (or existing model) is first converted to a starting template by stripping it down to only the required elements. Then a strategy for the elicitation session is chosen. This would likely be one of those discussed earlier: for example, one might choose to use a facilitator-driven discussion format following a reverse bow-tie approach or a stricter structured approach. As noted, 1-on-1s are more likely to be looser, more relaxed formats, but this need not always be the case.

We now come to the point in the workflow where we actually hold a workshop or 1-on-1. Regardless of whether these are conducted online or in person, there are several things that should be done. An introduction should be created for the session that covers the following:
\begin{enumerate}
    \item introductions of those attending;
    \item the project purpose;
    \item the purpose and scope of the CKBN being developed;
    \item the structure of the session;
    \item an overview of causal and Bayesian network modelling (if required);
    \item demonstration of the current version of the CKBN (including a qualitatively parameterised CKBN if appropriate);
    \item the narrative for the template model.
\end{enumerate}
In an online setting, sessions are usually shorter and expert time constraints may be more significant. In an hour long session, the facilitator should typically take no longer than 15 minutes to cover{\em all} of the introductory material . This is really only feasible for 1-on-1s or small groups (three or fewer experts). In larger group workshop settings, two hours is more likely to be the minimum for useful sessions. Covering introductory material adequately with new experts in either a short 1-on-1 or a group setting can be difficult; good preparatory materials sent beforehand can help the facilitator keep to time during the session.

The choice of software tool can prove critical to the success of the session. For the COVID case study, the team used the GeNIe software\footnote{Available from the BayesFusion website: \url{https://bayesfusion.com}.} as it provided the most fluid workflow for developing CKBN models with experts. GeNIe provides features such as renaming nodes directly on the canvas, easy ways to add and remove links, notes that can be easily added to the canvas or to nodes or to arrows (as well as to parameters). We noticed during the case study such live note-taking by the facilitator helped to engage and encourage the experts to contribute directly as model developers as opposed to solely a knowledge provider. Saving changes to file often is critical to avoid loss of elicited knowledge (but video recordings do mitigate potential loss). There is no natural way to do version tracking during elicitation in any of the software packages that our team regularly uses, although there is no impediment to setting this up to be triggered on file save. 

Each workshop or 1-on-1 should be recorded or transcribed, ideally both. Automated transcription can and should be used on recorded video files or live in an online video conferencing session. Transcriptions are useful as a reference to clarify later understanding and recall of key decisions made, and can also be useful if summarised with any key points extracted. During work on the CKBN structure, the facilitator should make annotations as often as possible on the nodes and arrows. The facilitator should make an effort to link these annotations to the transcripts, which can be achieved in most cases by simply using identical phrases in the annotation and the transcript (e.g., the facilitator can say the phrase from the annotation so that it is stored in the recording).

At the end of each session, the facilitator gives all attendees an overview of the next steps, and asks experts about their willingness to contribute to further sessions and activities if required. The facilitator should make sure to describe how the knowledge from the workshop will be incorporated into the CKBN, and how the CKBN will eventually support the applied BNs, models or other use cases.

Immediately after each session, all available members of the modelling team should conduct a short debrief discussion. The purpose of this discussion immediately after the workshop is to cross-check interpretations, consolidate understanding and clean up relevant documentation, while memories of the workshop are still fresh.

In the days following elicitation, the modellers will engage in several key tasks. These include:
\begin{enumerate}
    \item reviewing the outputs of the workshop and further refining the CKBN;
    \item ensuring elicited knowledge is documented correctly;
    \item ensuring causal mechanisms are complete and consistent;
    \item identifying areas for improvement in the CKBN or in the broader causal knowledge base; and
    \item further identifying the need and focus for follow-up sessions including 1-on-1s, workshops, surveys.
\end{enumerate}
Qualitative parameterisation may be applied to the CKBN if the structure is at a stage that is considered sufficiently stable to allow for it. If no further sessions with experts are expected, experts should be provided with follow-up updates on major changes to the CKBN structure via other means, such as via email. In the simplest case, this may just involve circulating a preliminary cleaned version of the CKBN structure and (if appropriate) the final version when it is considered complete.

Finally, and as always, the process is iterative. Whether the overall elicitation strategy continues to be considered suitable or not, it should be revisited in the light of knowledge gained from the recent elicitation sessions. This is not just based on the domain-dependent causal knowledge, but also knowledge about how well certain approaches worked, as well as any new requirements or ambiguities that need to be handled.

\section{Conclusion} \label{sec:discussion}

COVID-19 established the need for a rapid response in many areas of science, and created the right environment for modellers to experiment with methods for quickly developing models that captured evolving expert understanding of the new disease. This environment, combined with the constraint of initial limited access to good quality data, provided the incentive for our team to employ a staged, comprehensive approach to model development, yielding qualitative causal models of disease pathophysiology. What emerged was the CKE process we have described, a process for collecting and organising causal knowledge into a knowledge base, that can be used as a rich source of information to support development of application-specific models for prediction, decision making and intervention effect estimations.

Prior to our COVID-19 work, there were existing methods for developing application BNs (particularly, the KEBN approach) as well as methods for creating an ontology containing causal knowledge that could support development of a causal BN. Our approach draws upon these, but emphasises the development of a causal knowledge base consisting of standalone causal models (CKBNs), with potentially overlapping and competing features (for example, alternative hypotheses), that together with the top level framework that maps the relationship between the different CKBNs, form the causal knowledge base. A key advantage is that the causal knowledge base contains standalone models that are internally causally consistent, providing valid (whether or not true) causal accounts of elements of the domain from the very beginning. The causal knowledge base also goes beyond providing a list of isolated potential causal relationships --- it also provides the broader {\em context} for these causal relationships. This has several benefits: individual CKBNs can be shared and validated externally; they can be qualitatively parameterised to support validation of both CKBN structures and future application models; they can support communication with external stakeholders, as well as communication and organisation within the team; and they provide a database of vetted causal information that can be trusted as a source of information for the development of application models.

Being based on causal knowledge, a key component of the CKE necessarily involves engagement with experts, who can help in a number of ways, including by identifying the causal knowledge in existing models, data and literature, and providing their own knowledge, from firsthand experience, as well as their own inference and mental models of the problem. There are a wide variety of techniques for eliciting such causal knowledge from experts, and many existing approaches are manual and bespoke to the question at hand. The issue of how to tackle this kind of elicitation in a repeatable way is rarely discussed in the literature on BN applications. Recognising this, we have provided an approach based on the processes used during the COVID-19 model elicitation that provides flexibility for designing the elicitation process as needed, including the approach to scheduling workshops, 1-on-1s and employing surveys, developing elicitation materials and approaches to validation. There is still significant scope to make many aspects of this process more rigorous, and this is currently being pursued by members of our team, through the development of a structured group-based elicitation protocol for causal BN structures.

We are already applying the CKE approach in our other projects, including in modelling work on lower respiratory tract infection, cystic fibrosis, sepsis and serious infections as well as in domains outside of health. We have found being explicit about the divide between CKBNs and application models to be particularly beneficial, as it is extremely efficient to have a reusable store of causal knowledge that is large and detailed, while allowing the application model to use just what is needed to be lean, efficient and well-designed for the intended application. We would not suggest the CKE is suitable for all projects --- in particular, the overhead for a smaller project may not be warranted. Also, in cases where the purpose and scope of the application model is very sharply defined and experts are in good agreement, or there are good existing models that can be readily adapted, it may be preferable to work directly on the application model from the beginning. However, for those needing to build more than a few application BNs or who work regularly on the development of BNs for a particular domain, CKE can provide large efficiency and quality benefits.

\bibliography{main}

\newpage
\appendixtitles{yes} 
\appendixstart
\appendix
\input{appendix-techniques}

\end{document}

%% file: appendix-techniques.tex
\newpage
\section{Examples from the case study } \label{sec:toplevelframework}
\label{sec:examples}

\begin{figure}[h]
  \includegraphics[width=1\linewidth,page=1]{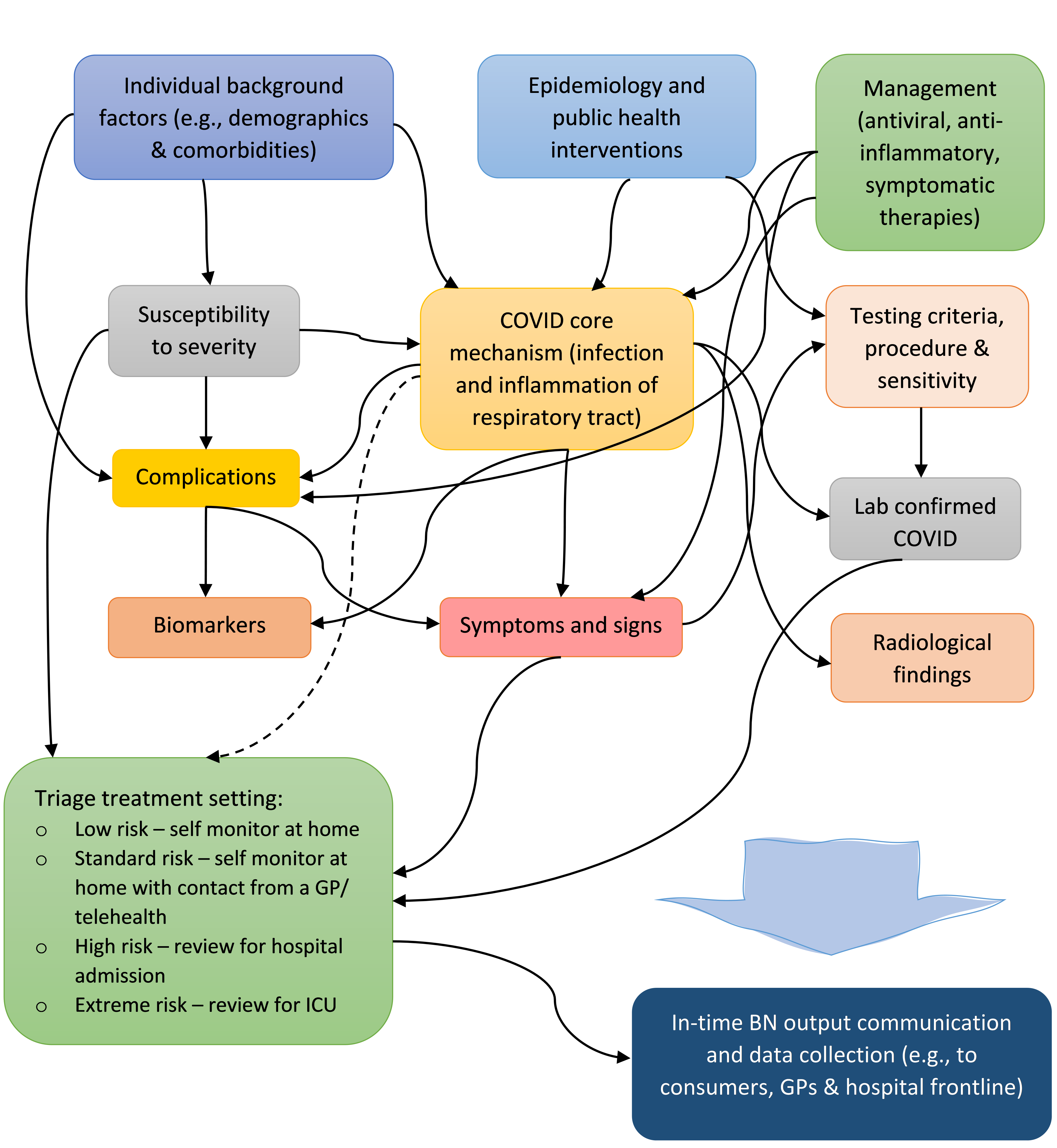}
  \caption{The top-level model for the COVID-19 case study. This was named and versioned as `Conceptual framework v1.1' in April 2020.}
  \label{fig:toplevel}
\end{figure}

\newpage
\begin{landscape}
\begin{table}[h!]
    \centering
    \scriptsize
\begin{tblr}{
  width = \linewidth,
  colspec = {Q[62]Q[300]Q[200]Q[85]Q[44]Q[212]},
}
\textbf{CKBN name} & \textbf{Description}                                                                                                                                                                             & \textbf{Development notes}                                                                                                                                                                                                                                                               & \textbf{Status}             & \textbf{Citations} & \textbf{Other notes}                                                                                                                                                                             \\
Respiratory        & Represents the initial
  pathophysiological process of SARS-CoV-2 in the respiratory system, including
  concurrent pathways down to key downstream complications such as multi-organ
  failure. & {~- Straw model\\
    ~- Expert 1-on-1 for workshop
  prep\\
    ~- Workshop (with starting
  template)\\
    ~- Expert revisions~ and validations via 1-on-1s\\~}                                                                                                                       & Expert validated, Published &                    &                                                                                                                                                                                                  \\
Complications      & Represents the
  physiological processes underlying the progression of COVID-19 from the
  pulmonary system to other organs.                                                                     & {
  
  ~- Starting template (based on
  simplified Respiratory BN)\\
    ~- Multiple workshops and
  1-on-1s\\
    ~- Further revisions to CKBN after
  development of application model (known as the Progression BN)\\
    ~- Final expert validation workshop
  prior to publication} & Expert validated, Published &                    &                                                                                                                                                                                                  \\
Background         & Captures~ factors affecting acquisition of a COVID-19
  infection.                                                                                                                               & {
  
  ~- Starting template (target node, plus
  list of potential background factors)\\
    ~- Workshop}                                                                                                                                                                                & Incomplete draft            &                    &                                                                                                                                                                                                  \\
Testing            & A comprehensive representation of the logistical factors affecting the
  sensitivity and specificity of a COVID-19 PCR diagnostic test.                                                          & {
  
  ~- Starting template (derived from
  Background BN)\\
    ~- Workshop\\
    ~- Series of 1-on-1s with individual
  domain expert}                                                                                                                                                 & Expert validated, Published &                    &                                                                                                                                                                                                  \\
Signs  Symptoms    & Represents the way in which signs and symptoms arise, based on the spread of
  infection to different sites, and the types of complications the infections
  gives rise to.                      & {
  
  ~- Starting template (containing sites
  of infection and known complications of the disease)\\
    ~- Workshop (linking in signs 
  symptoms to the starting template)}                                                                                                          & Draft                       &                    &                                                                                                                                                                                                  \\
Immune Response    & Representation of the local and
  systemic immune response to infection by SARS-CoV-2.                                                                                                           & {
  
  ~- Starting template (temporal/dynamic)
  \\
    ~- Workshop (with starting template)}                                                                                                                                                                                            & Multiple draft versions     &                    & {
  
  ~- Substantial revisions to template
  during initial workshop\\
    ~- Wide-ranging opinions on key
  processes from experts, several viable competing hypotheses and causal
  accounts} \\
Vascular           & Representation of the impacts to the vascular system from a SARS-CoV-2
  infection.                                                                                                              & {
  
  ~- Simplified Respiratory BN as
  starting model\\
    ~- Series of 1-on-1s with multiple
  experts}                                                                                                                                                                              & Expert validated            &                    & ~- Main conclusions incorporated back
  into Respiratory BN                                                                                                                                      
\end{tblr}
    \caption{A list of the CKBNs in the COVID-19 causal knowledge base. References for published CKBNs are included in the final column.}
    \label{tab:ckbn_list}
\end{table}
\end{landscape}

\newpage
\begin{figure}[h]
    \includegraphics[width=1\textwidth]{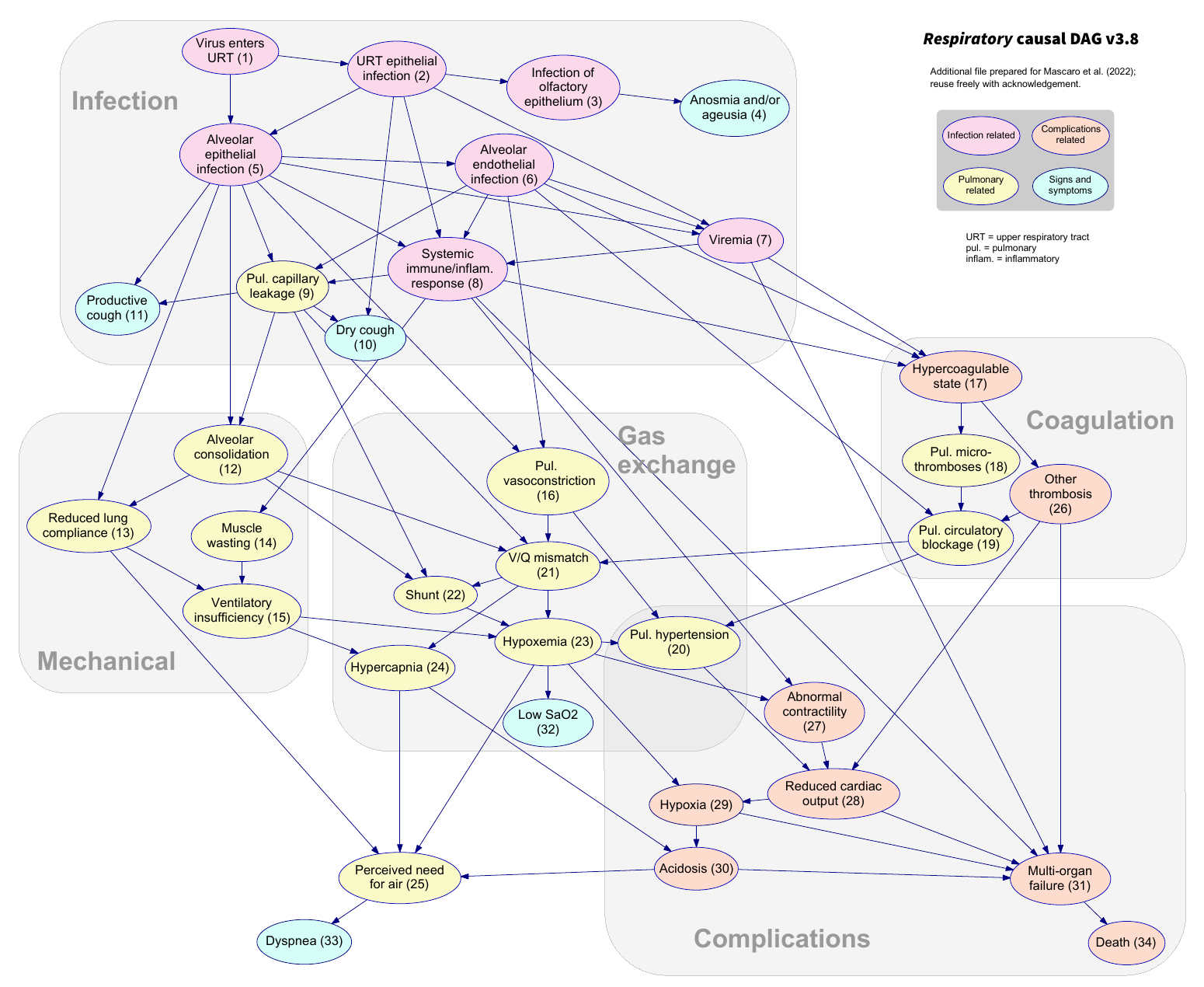}
    \caption{The structure for the Respiratory CKBN. (Respiratory causal DAG v3.8.) This depicts the initial pathophysiological process of SARS-CoV-2 in the respiratory system, outlining multiple and often concurrent pathways from viral infection to key downstream complications such as multi-organ failure. Some variables are latent (i.e., not directly observable) but their probability distributions can be inferred from observable evidence such as clinical signs, symptoms and laboratory measurements, not all of which are shown in the diagram. Many mechanisms described in the BN can be influenced by background factors such as age, sex, and comorbidities, which are also not shown. BNs are acyclic, so feedback loops that may occur as the disease progresses are not included in the diagram. We divide the nodes into four color-coded categories: Infection process (pink), Pulmonary details (yellow), resulting Complications (orange), and a few illustrative examples of Signs and symptoms (cyan). Within the pulmonary system, we distinguish (using background boxes) three pathways from Infection to possible Complications: involving problems with Mechanical operation of the lungs, Gas exchange, and Coagulation. \textit{(Figure and caption reproduced and revised from \citet{mascaro+2023}.)}} \label{fig:respCKBN}
\end{figure}

\newpage
\begin{figure}[h]
    \includegraphics[width=1\textwidth]{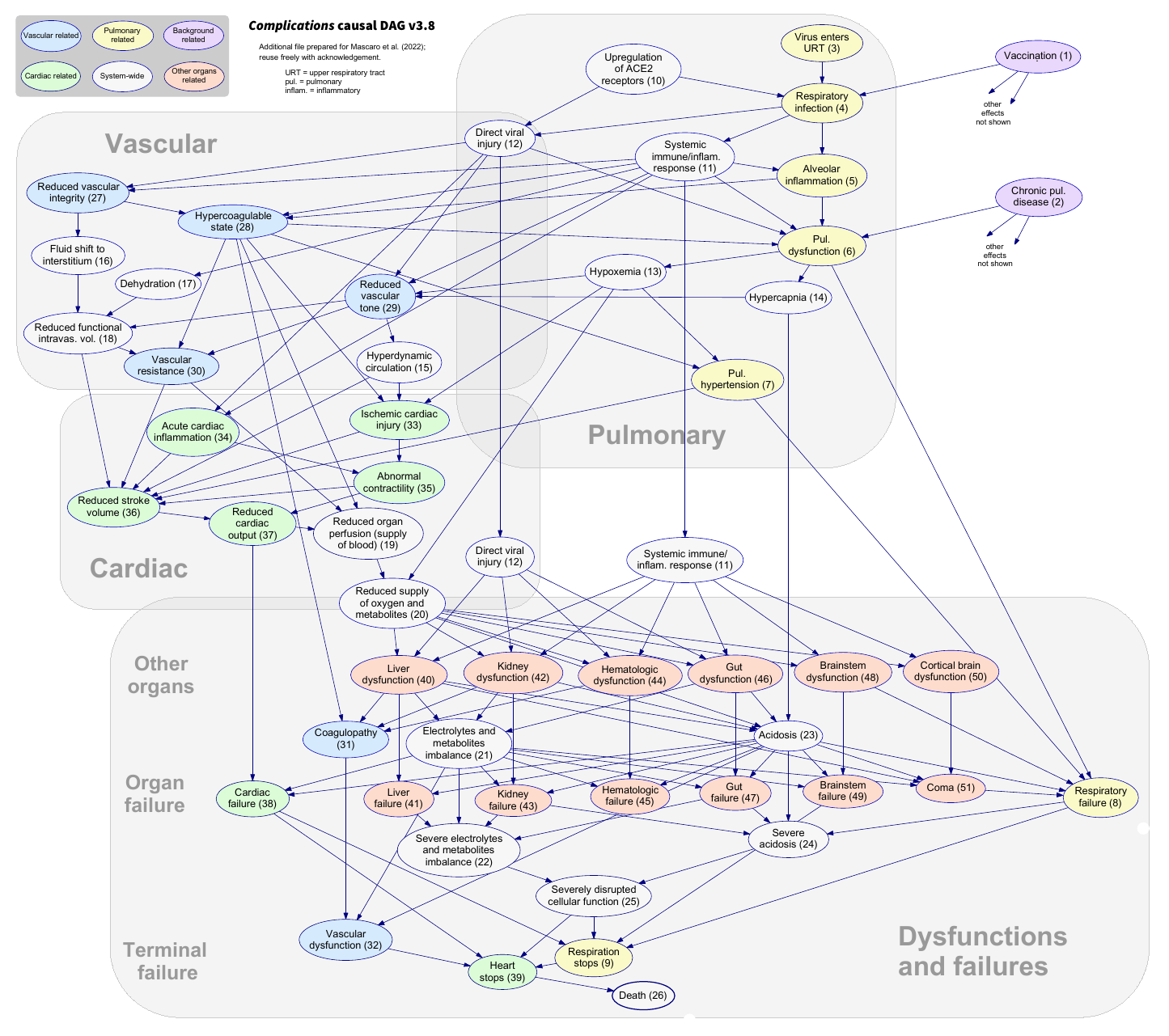}
    \caption{The structure for the Complications CKBN. (Complications causal DAG v3.8.) This depicts the main physiological processes underlying the progression of COVID-19 from the initial infection in the Pulmonary system (nodes in yellow) to complications in other organs. Vascular (nodes in blue) and Cardiac (nodes in green) systems are modeled in more detail due to their likely earlier involvement and greater system-wide impact on Other Organs (nodes in orange), i.e., liver, kidney, hematologic, gastrointestinal, cortical and brainstem dysfunction. Mechanisms that have a system-wide impact are colored in off-white, and we include two illustrative examples of background factors (nodes in purple). \textit{(Figure and caption reproduced and revised from \citet{mascaro+2023}.)}} \label{fig:respCKBN}
\end{figure}

\newpage
\begin{figure}[h]
    \includegraphics[width=1\textwidth]{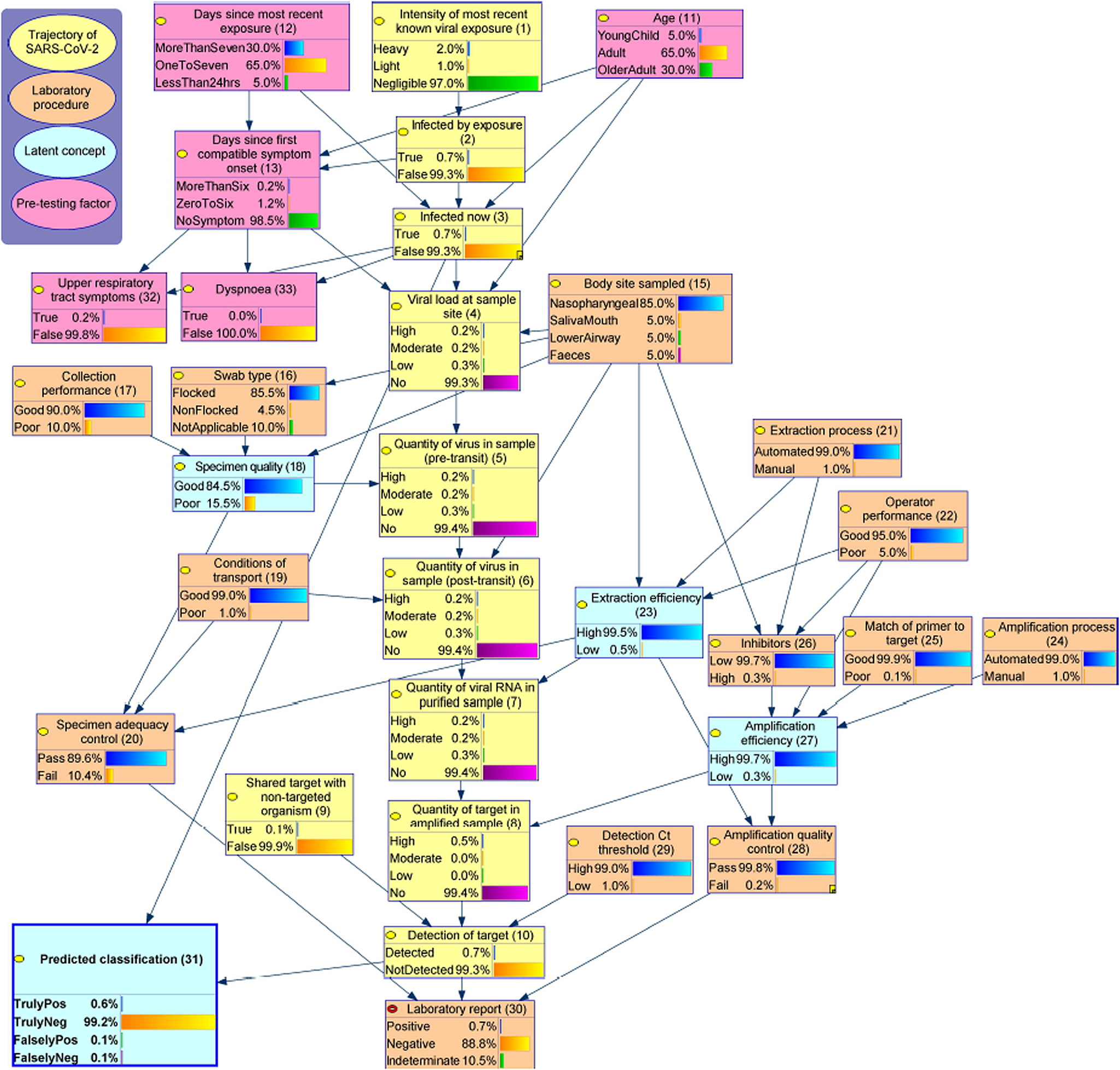}
    \caption{The structure and posterior beliefs (based on the published qualitative parameterisation) for the Testing CKBN, a causal BN of RT-PCR testing of SARS-CoV-2. This diagram presents the model structure, variable values and marginal distributions (i.e. when nothing is known, other than that a test has been conducted). Appendix A of \citet{wu+2021} provides a comprehensive variable dictionary for this model. Detailed conditional probability tables can be accessed via Appendix B\_1 of the same paper. \textit{(Figure and caption reproduced and revised from \citet{wu+2021}.)}} \label{fig:respCKBN}
\end{figure}

\newpage
\section{Techniques for developing CKBNs} \label{sec:building}

It is difficult to provide a clear and rigorous workflow for the development of a CKBN. As with the causal knowledge base itself (Section~\ref{sec:ckbpurpose}), developing an individual CKBN is a very open, creative and subjective process, and what works well for one problem, modelling team and expert group may not work so well for another.\footnote{We wish to emphasise again that whether or not the purpose and scope have been achieved is {\em not} subjective; but the exact form of the CKBN is under-determined by the purpose and scope, and may be used in a number of ways beyond the project, and hence the form of a CKBN is subjective. See again Figure~\ref{fig:multi_multi_organ} and associated discussion.} Rather than give a prescriptive workflow that is of little practical use, or otherwise give up on the goal of providing any guidance entirely, we can provide useful guidance by collecting together the techniques that we found helpful during the COVID-19 modelling effort (and in some cases, other work), along with some recommendations on which techniques work well or not so well under what circumstances.

We catalogue the techniques that we have used in the COVID-19 case study below,
describing where and how we have used them. We divide these techniques into 3 different categories that approximately correspond with the main phases of CKBN development: utilising causal knowledge sources, developing the CKBN structure, and checking and revising the CKBN.

\subsection{Utilising causal knowledge sources}

Figure~\ref{fig:ontology} divides causal knowledge sources into literature, models, experts and data. In addition, we noted inferred knowledge. This is, of course, inferred from the other knowledge sources, but there are nonetheless inference-specific techniques that are worth discussing.

\subsubsection{Literature}

The ways of obtaining causal knowledge from literature are exceptionally broad and varied, and typically unsystematic. We describe a small number of general techniques here to provide modellers an overview of some key options that affect the way in which they may approach the literature.

However relevant literature might be discovered, it should be tied to the causal BN by documenting it in the variable dictionary for the BN. References can support both nodes (such as sources for variable definitions) and arcs (such as sources supporting or undermining the existence of a causal relationship, or associative relationship).

\paragraph{General research} Any of the techniques that may be applied to conducting research in general can also be applied to discovering causal knowledge for a BN in the knowledge base. Using the purpose and scope of the causal BN, one can generate a set of research questions, search keywords and terms, relevant journals and publications as well as identify relevant researchers or other experts who might write papers and articles or produce other material related to the causal BN's purpose. The key point here is that discovering and organising sources of causal knowledge for a BN can utilise the very same process as any other form of research.

\underline{Case study:}  Our team collected references into a Zotero database that could be used for any of the modelling team's subsequent work. We also stored key causal claims and implications made by a reference alongside it in the reference database. While this was done manually, a structured system could be used to report a list of causal claims, which can then be used to support, compare or integrate knowledge in a causal BN.

\paragraph{Systematic review} A systematic research process supports reproducibility, comprehensiveness and diminishes the impact of bias. For causal BNs, this would involve making the purpose and research goal explicit and clear from the beginning, documenting the search methods used (such as keyword searches, snowballing, expert consultation) and planning and documenting the way references will be included or excluded as sources.

There are significant drawbacks to using systematic reviews to support development of a causal BN. The effort required to do it well can be considerable. In addition, the process does not fit well with the iterative development model for causal BN development (a method we recommend). In particular, starting small and growing the model as the need arises and in the manner best suited to that development, does not work well with specifying all one's methods as clearly as possible in advance. A compromise may be to specify what one can in advance (e.g., purpose, initial search methods, initial inclusion and exclusion criteria) and document any changes along with how individual references were found.

\paragraph{Elicit literature from experts} Experts can and ideally should support modellers in identifying and assessing relevant literature. This can be done in a structured way, by eliciting relevant literature via surveys or as part of workshops or 1-on-1s, via reviews and revisions of variable dictionaries, or in an unstructured manner through informal discussions.

\underline{Case study:} We  often requested relevant literature from experts via email prior to an upcoming workshop or 1-on-1, as well as less formally during and at the end of workshops and 1-on-1s. Experts also assisted more directly by reviewing and suggesting references in variable dictionaries for the causal BNs. 

\paragraph{Post hoc confirmation and disconfirmation} In many cases, a model (or large parts of it) may be built without any explicit usage of literature. This can make it much easier to develop a causal BN, and is often the only feasible way for a group workshop to be conducted. (See in particular the techniques for model development described in Section~\ref{sec:devCausalBnStructure}.) Literature can be sourced after such models have been developed, both to check the validity of the structure developed and aid in its revision, as well as to document the ways in which it is supported or undermined by the literature.

Even assuming a group workshop is conducted in a structured and unbiased way (itself quite difficult for structure elicitation), searching for sources that justify causal claims by the elicited model runs the strong risk of confirmation bias. Unfortunately, this is a difficult problem to solve, as the literature itself is biased towards publishing positive results, rather than negative and null results.\footnote{Hence the lack of any published negative or null result does not often strengthen the case of a positive result.} We do not have a good solution, however documentation can help, as sources can be checked by external parties (such as peer reviewers) and contradicting sources more readily found.

\paragraph{Literature monitoring} Where an application model may be operationally deployed in a high consequence setting (such as for clinical care), it may be worthwhile ensuring the causal knowledge base that supports the application model is up-to-date via literature monitoring. This is less likely to be of value in problem domains that can be treated as slow-moving or static for the duration of the analysis. 

\underline{Case study:} Research output and perspectives on COVID-19 grew and changed rapidly over time. Hence, the team actively monitored journals, reviews, news sources and social media for new perspectives and understanding on the pathophysiology of COVID-19.

\paragraph{Causal claim extraction} This is a more formal approach to deriving causal knowledge from the literature, which analyses the causal claims by individual papers. A key advantage is that it can produce a more comprehensive causal knowledge base.

\underline{Case study:} Our team made attempts at a manual process for extracting causal claims from research papers. This was an intensive process, and despite success in generating informative BNs from individual papers, most of the causal knowledge extracted was of little new value to the overall modelling process. However, we believe there may be opportunities for using automated natural language processing tools to extract causal claims that may make this approach more feasible; or otherwise simplified approaches to manual extraction, such as extracting only key new claims, or claims referenced in abstracts.

\paragraph{Generative artificial intelligence} With the recent rapid advance in the capability of large language models (LLMs), the team has been exploring the use of these tools as assistants in the development of causal models. While LLMs currently exhibit weak reasoning (and causal reasoning) capabilities, they can perform reasonably well in the role of another kind of expert~\cite{}. The key advantage over human experts is the rapidity and ever present access of this expert for consultation by the modellers. However, any knowledge derived from an LLM ultimately has to be well supported independently, as the risk of plausible sounding but entirely false information is still extremely high. 

\subsubsection{Models}

Existing causal models (whether BNs, DAGs or other forms of causal model) and potentially conceptual models can be incorporated into the causal knowledge base either directly, or by supporting the development of knowledge base-specific causal BNs. In our experience, reuse of causal BNs across different research groups is relatively uncommon. This might be because each modelling problem tends to have scoping and requirements that are very particular to that problem; it may be because the kinds of data available are different to each group; or it may also be because each research group has a different way of organising their knowledge that works well for the experience, purposes and tools of that particular group.

With this in mind, we believe there are two important types of source models: causal and conceptual models that have been developed within the team, and those that have been developed by others and that may be publicly available. Either can be adopted and adapted directly, but more likely than not, would be used as a source of causal knowledge that can be incorporated into the knowledge base selectively.

\underline{Case study:} We drew upon the causal knowledge contained in our team's earlier models on pneumonia and cystic fibrosis to develop an initial tentative picture of potential COVID-19 pathophysiology. This was quickly revised as we talked with experts, and better understood how the disease differed from the respiratory issues encountered in other diseases. While we did not make direct use of causal knowledge from other models, we did use concepts such as the neutrophil to lymphocyte ratio that was found to be key in other predictive models.


\subsubsection{Experts} \label{sec:experts}

Eliciting knowledge from experts is an important topic, and much has been written about it. Somewhat less has been written about elicitation of causal knowledge and causal structure. Here, we just provide a brief outline of some of the key techniques that modellers should consider. In Section~\ref{sec:elicit}, we provide a relatively detailed description and workflow of the process we used during the COVID-19 work to elicit causal knowledge before, during and after group workshops. 

\paragraph{Structured versus unstructured elicitation} Before describing the techniques, one factor affecting almost any choice of technique is how structured the technique is. The process may be carefully guided, with an emphasis on standardised questions, or more loosely guided, allowing for more open and creative responses and exploration. Advantages of a strongly structured process are that it improves repeatability, significantly reduces the chance of cognitive biases distorting the causal knowledge elicited and recorded, makes it easier to check that the elicitation has been comprehensive and allows more opportunities for validation of the elicited knowledge~\cite{}. Unfortunately, for causal knowledge elicitation specifically, the disadvantages can be significant. In particular, causal knowledge elicitation often depends on substantial iteration on a complex structure, and current structured processes (such as those inspired by Delphi techniques) do not deal well with either significant iteration or complex structures. As such, a strongly structured process can be extremely burdensome for non-trivial problems and it can significantly stifle and constrain expert input. To make the process practicable, modellers often have to make substantial modelling assumptions before talking to experts to make a structured process work, which can lead to significant issues. 

\underline{Case study:} We had limited success with our attempts at the more structured forms of elicitation and hence our approach was primarily unstructured. Generally, we found experts were uncomfortable enough with the assumptions in our structured attempts that it made elicitation difficult, even when those processes were designed with the aid of other experts. For example, we developed a temporal structure template for immune system changes, which we checked with one expert, but found in the workshop that the temporal assumptions made it difficult to model events correctly, and had to abandon it early. 

\paragraph{Surveys} Surveys are a good way of communicating asynchronously with experts between workshops and meetings, particularly for things where experts may need time to reflect on or research the answer to a particular problem. Surveys that lead into a workshop also serve as a way of ensuring the expert knows what to expect. It is important to ensure, however, that surveys are not burdensome, and are not so open-ended that experts may feel obliged to spend too much time on them.

\underline{Case study:} Prior to the first workshop, experts on an open mailing list were sent a relatively large survey (with 95 questions) that attempted to elicit causal knowledge and assumptions, as well as some estimates of causal strengths. Responding to the survey was optional. While we had good engagement, this was likely due to the interest in the very new and notable disease, and many experts noted their struggle to provide answers to questions due to disagreements underlying the assumptions in the survey.

In most cases, we forewent surveys for causal knowledge elicitation and instead provided materials (example models and references) by way of preparation.
Later, particularly during parameterisation and evaluation, the lead-in surveys that were sent by the team were kept relatively short, and designed to elicit small amounts of information and occasionally to provide guidance on what to expect in an upcoming workshop.

\paragraph{Informal discussion} While most methods of obtaining causal knowledge from experts involves some form of prepared and organised elicitation, there are often cases in which informal chats are more conducive to the sharing and recall of information. It can therefore be useful to provide opportunities to discuss the modelling problem less formally. This can be done via a number of means and mediums, either synchronously in person, by phone or by video call, or asynchronously via email, chat and instant messaging.

\underline{Case study:} The team explored a number of options for encouraging general discussion via informal channels, including via Microsoft Teams channels, email discussions (both with larger and smaller groups) and 1-on-1 video calls. Email discussions and particularly 1-on-1s proved the most fruitful (discussed next).

\paragraph{1-on-1s} A 1-on-1 (or small scale) elicitation eschews the group setting to focus on a discussion with one individual expert, or in some cases 2-3 individuals who work closely together. There is a single `facilitator' (acting as an interviewer) often accompanied by others from the modelling team who may also participate in a number of ways. While this does mean a 1-on-1 can involve quite a few more than 2 people, there are several characteristic factors that distinguish it from group elicitation and also make this name more suitable than alternatives:\begin{itemize}
    \item A 1-on-1 aims to allow a particular expert perspective to be explored in depth. Even if there are multiple experts, they should be selected such that they {\em overlap} substantially in terms perspective, ideas and values, enough that disagreements are very likely to be minor and very unlikely to impede discussion. Hence, the goal is to explore a single mostly cohesive perspective.
    \item When multiple experts are selected for a single 1-on-1 session, they should be very comfortable with each other (e.g., work together very regularly), so as to minimise dominance issues, and foster an accepting environment in which to share ideas, again allowing the experts to act cohesively.
    \item While there may be multiple people from the modelling team, the primary facilitator drives the discussion. The modelling team should agree on the general direction of the 1-on-1 ahead of time, and should act cohesively, to make the expert(s) feel at ease, provide the expert with confidence in the process, and hence give the best chance of the expert sharing the best information they can.
    \item In a Delphi process, participants provide their responses independently in a first round to avoid an array of biases
. It is difficult to use a Delphi-like process directly to elicit causal knowledge due to the need for significant assistance from the modellers; a 1-on-1 retains many of the benefits of the anonymous Delphi-like process, while providing experts with the assistance they need to contribute their causal knowledge.
\end{itemize}

A 1-on-1 is best approached as a relatively informal interaction to make experts feel comfortable, but it certainly involves preparation and should still include an overall structure and elements of formality. For example, for new experts, a brief outline of the project, purpose and modelling problem at the beginning of the session is essential. With any expert, a discussion of next steps and followups at the end are also generally advisable. It is also important to have a set of questions chosen ahead of time by the modellers, and these might be used to initiate and guide some of the discussion. A more structured process (such as working through a fixed survey) can work if it does not stifle in-depth discussions.

\underline{Case study:} The team utilised 1-on-1s extensively after the first set of group workshops
to validate the elicited models, and to explore areas of the models that required deeper examination. This provided ample opportunity for the modelling team to clarify ideas, intentions and definitions and to test that the models were able to accommodate the perspectives of individual experts (or expert groups). They also provided opportunities to elaborate on the model structures and hypotheses in a cohesive way, and avoided some of the pressure for experts to agree for expediency or politeness, which can occur within the setting of a group workshop.

\paragraph{Group workshops} The focal point of an elicitation process is often the group workshop. In general, group workshops for elicitation of causal knowledge can comfortably accommodate anywhere from 3-8 experts over the course of a single 2-4 hour session, either in person or remotely. They are an excellent, and sometimes the only feasible, way of obtaining a broad range of inputs, advice and oversight of the causal modelling problem when both expert and modeller time is limited, or when there are limited opportunities to iterate with experts.

While there are a number of excellent approaches to group elicitation for well-defined individual questions (or sets of questions), these approaches break down when applied to causal knowledge elicitation. Some of the reasons that these approaches fail for causal knowledge are as follows:

\begin{itemize}
    \item The number of possible causal arcs (without preventing cycles) between $n$ variables is $n(n-1)$, and the number of possible directed graphs over $n$ variables is $2^{n(n-1)}$.\footnote{To give a sense of the numbers here, a modest 20 variable model has 380 possible arcs. It also has $2^{380}$ directed graphs but fewer DAGs --- approximately $2^{240.4}$.} Even if the variable list is fixed and each arc is considered individually, rather than the complete impossibility of considering every possible full DAG, being comprehensive is infeasible.
    \item Total and partial orderings over the variables can help reduce the number of arcs that subsequently need to be considered. However, there are a larger number of total orderings ($n!$) than there are possible individual causal arcs (but not possible DAGs), so total orderings need to be based on some easily accessible characteristic (e.g., if the timing is already well-known) to be helpful. Partial orderings (often in the form of tiers) are typically more useful, as it is much easier to group variables based on their approximate timing or order, and then sort those groups.
    \item The above grim numbers {\em assume the variables are known}. The greatest difficulty is in identifying variables --- and not just known measurable variables relative to some dataset(s), but latent variables that mediate the causal processes, along with variables that may be measurable but are currently not measured.
    \item The process above is necessarily {\em iterative}. As new variables and causal relationships are identified, this often highlights the need for further variables and additions or revisions to the set of causal relationships.
\end{itemize}

While there is no hope that such a process can be comprehensive, in practice, there are substantial and systematic independencies in the causal structure that make the overall process manageable. Navigating and incorporating these into the model is still mostly an art. We discuss a set of approaches that we used during the COVID-19 modelling to navigate this process in Section~\ref{sec:elicit}.

\subsubsection{Data}

Data is generally distinguished from information, and both in turn from knowledge \cite{}. Data is considered to be the closest of the three to raw and symbolic, and the furthest from meaningful and useful.\footnote{Since we are referencing a definition of data, it is worth noting that we treat `data' as a mass noun as it is more natural for our purposes; e.g., big data, processed data, meaningful data, valuable data refers to properties of the mass of data, not the properties of each individual datum.} Information is data that has been processed and has some context and meaning. Knowledge is generally information that belongs to a well-integrated and well-tested set of information. Data in its most raw form is not generally of much direct use for a causal knowledge base. However, processed, analysed and contextualised data can contain meaning, information and knowledge that can be integrated into a causal knowledge base.

\paragraph{Data dictionaries} A dataset can in principle be produced by anything capable of producing symbols. Typically, however, data is collected according to a designed and intentional process --- and, importantly for us here, the design and intent are based on a model (often implicit) of the process that produced the data. Data dictionaries are extremely common artefacts associated with datasets that provide insight into what these models are. Large scale database systems, and standardised data collection models, aim to make these models more explicit, and often go by the name `data models' in this context \cite{}. Data models rarely (if ever) require {\em causal knowledge} to be captured as part of the model. Nonetheless, causality frequently leaves its fingerprints plainly exposed in a well-specified data model. Hence, extracting causal knowledge from data dictionaries is often valuable and worthwhile.

\underline{Case study:} The team had access to several data dictionaries early on in the process (particularly, from the ISARIC and LEOSS groups), for datasets that we were expecting to be available in short time. As it happened, the team's access to the datasets themselves were delayed, and while waiting, we examined the data dictionaries after the initial causal models were produced to identify matching variables. While not intentional at the time, this process helped the modellers identify how measurable variables were causally linked (not just associated) to key latent causal mechanisms in COVID-19.

\paragraph{Data analyses} With access to data generated from the process of interest (or from some analogous process), there are various ways to work from the data back to the generating process. When this is done with primarily automated methods, this is called causal discovery \cite{}. When statistical analyses are used, this is often (today) called causal inference \cite{}. However, data analyses can shed light on the causal process in a number of other ways. This includes visualisations, summary tables, training and exploring some subset of hypothesised models (whether causal models or otherwise) and individual case analysis. Transformation and data processing is often treated as an input into the analysis process, but they are also often integral to the causal knowledge obtained in such analyses.

\underline{Case study:} Data analysis was a significant component of the project. Initial data analysis was performed for the LEOSS dataset, without direct access to the dataset itself. Instead, analysis scripts were to the LEOSS team to be run, and the resultant summaries returned. This process highlighted for us just how significant the implicit modelling assumptions around data could be. In particular, in addition to the causal hypotheses around the COVID-19 disease itself that we were attempting to test, there were also causal processes around data collection and recording that were mostly unknown to us. This suggested possibly pathways to improve these kinds of arms-length analyses.

We subsequently obtained direct access to the ISARIC dataset via the IDDO group. In this case, we performed an array of analyses, including on missingness, mutual information and correlations, tables and visualisations of marginal (i.e., one dimensional) distributions, as well as tables and visualisations of associations amongst 2 or more variables. The higher dimensional analyses in particular were aimed at identifying not just associations but also at generating causal knowledge either directly from the analysis, or indirectly via leads that the modellers could investigate.

\paragraph{Causal discovery} Causal discovery attempts to discover causal relationships from either data alone, or data coupled with expert knowledge \cite{}. (As noted, data almost always has some in-built assumptions underlying its formatting and hence there is always some degree of expert knowledge involved.) For our purposes here, causal discovery is another kind of data analysis that is highly automated. It promises to identify or support causal knowledge generation at scale, and make the process of building a causal knowledge base much more efficient. There are several different approaches, and these may differ in their suitability for a particular task.

\underline{Case study:} The team explored the use of causal discovery (making use of the PC and CaMML approaches \cite{}) later in the project, particularly as a way to identify relationships that may have been missing in the primarily expert informed models, but also as a way to test the causal relationships that had been included in the models. These were applied late in the project, and hence typically applied to our application model, rather than our causal knowledge base models directly. For the PC algorithm, individual arcs were tested one at a time, essentially taking advantage of its basis in statistical tests directed towards causal inference to provide evidence for the presence of individual causal arcs. For CaMML, the model as a whole was tested against the data, to see how much of a change CaMML would suggest from the hypothesised causal structure.

\subsubsection{Inference \& logic} 

Inference and logic can be coupled with other causal knowledge to generate new causal knowledge that is difficult or impractical to obtain from other sources. We describe just a small sample of these techniques that are frequently useful.

\paragraph{Logical model requirements} When developing models with input from experts, literature and often data, it is frequently the case that only part of the causal story is provided. A partial causal story may be adequate for the purpose and scope; if not (or if unsure), the causal model can be examined for places in which alternatives are feasible. If an effect can be explained by multiple causes, it is worth ensuring that either all notable causes are included within the causal model, or that it is clear that alternative causes will be handled by adding noise to the child distribution (and any other distributions as relevant). A Noisy OR is an example of a local structure that requires any other alternative causes to be handled explicitly, via its `leak' parameter. This pattern of accounting for logical alternatives, however, is not just limited to Noisy ORs, and applies to any parent nodes (direct causes) to an effect, as well as to more ancestral causes.

Nodes can also be checked to see whether they are mutually exclusive. It is {\em not} a requirement that nodes be mutually exclusive, particularly if the overlap is recognised and intentional --- for example, it can be perfectly fine to have a 5 state node that is deterministically grouped into a 2 state node, or to have two nodes representing partially overlapping concepts in a causal model because that is the way they are known and referred to by experts or literature. However, examining a model for unintentional cases of overlap can help identify where a model needs further work, or may generate causal knowledge more directly in the process of separating out the overlapping concepts.

\underline{Case study:} In the Respiratory BN \cite{}, there are three pathways that may lead to more serious outcomes from a COVID-19 infection. The `mechanical' pathway was identified as an alternative pathway to the other two during the process of elicitation that could not be accounted for by the mechanisms on the other pathways. In addition, it was recognised that shunting and V/Q mismatch were closely related mechanisms, in which shunting is essentially an extreme form of V/Q mismatch; however, the two mechanisms were retained in the model as separate nodes, since some interventions were potentially specific to shunting.

\paragraph{Analogy} When a causal process is not yet very well understood and little direct information is available for it, it can be useful to examine similar processes (and models of similar processes) in case they contain usefully similar mechanisms. When building a causal knowledge base, the goal is not necessarily to {\em validate} these analogical processes, but merely to document them as {\em possibly} relevant. Ideally, models and pathways based on the analogy are not merely adopted, but also adapted, shaping them into their most plausible forms for the causal process of interest.

In reality, analogy is an inherent part of {\em any} modelling process --- here, the emphasis is on conscious attempts to reuse ideas, models and pathways from elsewhere to help model the causal process of interest.

\underline{Case study:} In the team's initial version of the Respiratory BN, the main (gas exchange) pathway was based predominantly on an existing model of lower respiratory tract infections. While this model was revised heavily in discussion with experts due to the differences with COVID-19, the same process of analogy was used to develop other pathways in the Respiratory BN, including the (revised) gas exchange, thrombotic and mechanical pathways. The Complications BN made use of COVID-specific knowledge around initial infection, but was driven heavily by the analogy with complications arising from other serious infections.


\subsection{Developing CKBN structures} \label{sec:devCausalBnStructure}

In this section and Section~\ref{sec:checkBn}, we focus on individual graphical (and particularly causal DAG-oriented) techniques to developing and revising CKBN structures, separate from any particular knowledge source, development context or purpose. In Section~\ref{sec:elicit}, we turn our attention to a broader end-to-end process of eliciting a CKBN for the knowledge base. 

\paragraph{Target nodes} A key technique that can be used on its own, but often forms a basis for other techniques, is to focus development of the structure around target nodes; i.e., the key variables that are likely required to fulfil the CKBN's purpose. For many modelling problems (including all of the CKBNs for COVID-19), there is often no more than a few target nodes, and the value of other nodes lies in the degree to which they can inform or affect the target nodes. Target nodes are often identified by the modellers with the help of the advisory expert prior to engagement with other knowledge sources, including other experts, literature and data.

Once the target nodes are established, one can work outwards to other relevant causes and effects, and/or identify pathways between the target nodes. This can be done in more or less orderly ways, and some of the more orderly methods are described below.

\paragraph{Bow-tie elicitation} Bow-tie analysis \cite{} is a method for performing risk analysis that evolved separately from analysis and elicitation methods for BNs based on representing causes with logic gates. There have been some more recent efforts to map these techniques directly into BNs \cite{}, though generally the emphasis has been on preserving and extending the logic gate approach, rather than applying it as a BN elicitation technique in its own right.

Bow-tie analysis begins with an event of interest, known as a critical event (typically a fault or hazard or something undesirable), that works much like a single target node as described above. There are then two parts to the bow-tie graph. One part spreads out back in time (or to the left) to identify what (or what conditions) can cause the critical event. This type of analysis is called fault tree analysis \cite{}, and has a relatively long history in reliability engineering. The second part spreads out forward in time (or to the right) to identify the consequences that flow from the critical event. This second type of analysis is called event tree analysis \cite{}. The critical event coupled with the fault tree and the event tree resemble a bow-tie, hence the name.

The key value for CKBNs (and BNs more broadly) of these methods is in providing a useful technique and guide for developing CKBN structures that suit certain common problems well. For example, in a CKBN we frequently want to work back from something like the `critical event' --- more generally, some focal event of particular interest --- to its causes. We can do so in stages, asking for the most direct causes of the focal event, then checking if each proposed cause is indeed direct, or may operate mostly via some other cause or set of causes; and then moving back to the next layer of causes and so on. There is also value in identifying how causes may interact at a qualitative level to produce the focal event, such as whether there is an OR or AND relationship (in the case of BNs, using Noisy ORs or ANDs), or otherwise some other similar type of simple noisily-deterministic relationship.

Identifying the consequences of the focal event may work in a similarly staged fashion, working out the most direct consequences first, and then moving on to the more indirect consequences. If one expands {\em outward} in the form of a tree, each node would have only a single parent, and there would be no interesting interactions amongst the consequences. There is no restriction to trees (particularly in a BN), so interactions can still occur, however they are likely to be identified more haphazardly than when working backwards. This is because we try to determine all the parents at once when working backwards, while there is no necessary point at which all parents of a consequence are considered together when working forwards.

\underline{Case study:} We did not regularly make direct use of the bow-tie technique (instead working with the modified version described below). However, we did explore the right side of the bow-tie in one of the diagnosis models, identifying the signs and symptoms that would result from infection by a particular causative pathogen (either of SARS-CoV-2 or of other infections like influenza or bacteria) along with some of their interactions. 

\paragraph{Reverse bow-tie} During the COVID work, the technique we used most frequently was in fact a variation of the bow-tie method, which we will call a reverse bow-tie. In particular, instead of choosing a single focal event and working back to causes, and forward to consequences, the reverse bow-tie begins with two focal events, the start and end events (or more generally, a set of start and end events). In the reverse bow-tie, the goal is to elaborate down from the start event until the effects overlap with the end event, and/or likewise to elaborate up from the end events towards the start events. When using this method during elicitation, experts are allowed the liberty of suggesting any cause (effect), even when it may not be immediately apparent whether it will lead towards the start (end) event. While there is the risk that experts that may suggest events that are not relevant to the way start and end events are connected, experts do have in mind at all times what the start and end events are, which is typically sufficient, gentle guidance to avoid suggesting events that are very unlikely to be relevant to the model purpose.

Both the bow-tie and reverse bow-tie methods allow for indirect causal pathways between any of the nodes, such as via confounders (that is, common causes) and colliders (assuming evidence might be set on those colliders, or selection bias in the context of data trained models) (for discussion, see \cite{}). The reverse bow-tie method is better suited than the bow-tie method to identifying these indirect causal pathways in general, particularly between the focal start and end events.

\underline{Case study:} As noted below, our initial attempts to work with starting models (based on previous work for other infectious diseases) proved difficult. In those early efforts, sometimes all that remained after discussion of a starting model were a small set of start and end nodes. The level of disagreement was due in large part to our experts identifying that COVID-19 acted quite differently on the human body as compared to other diseases. Settling on the smaller set of start and end focal nodes led to fruitful outcomes for the Respiratory, Complications and Testing BNs (the latter of which developed as an offshoot of the general diagnosis BN). 

\paragraph{Expanding mediating pathways} This method is quite similar to the reverse bow-tie, however it is a little more demanding. The start and end nodes are chosen, and a causal arc is drawn between them. At each step, an arc is replaced with one or more pathways containing one or more mediating nodes. If the specified pathways between two nodes are not considered to account for all relevant pathways, either additional pathways are added, or the original arc is restored to deal with this `leak' (analogous to the leak in a Noisy OR). An advantage of this technique is that it ensures explicit checks are made for missing key pathways, but it can easily lead to dense BNs if the threshold for considering all pathways to be covered is too high.

\underline{Case study:} The method was not trialled with experts, but instead trialled with a set of small data-dependent modeller-developed BNs that were inspired by the larger expert elicited models. In the context of these smaller models (of no more than 6 nodes), the method worked well to identify problematic pathways, and eventually lead to the re-inclusion of Age as a direct cause of a patient's final Status.

While not trialled as a part of the COVID project, it is worth noting that this method can be extended to handle pathways with confounders and colliders between the start and end nodes, making it a good approach for causal mediation analysis in a CKBN or BN more generally \cite{}.

\paragraph{Starting models} If modellers are working in a domain in which they have already created a number of models, it is frequently the case that new modelling problems will need models that have many features in common with one or more of the existing models. In such cases, the modellers (perhaps with the advisory expert) can quickly identify features from existing models that would be of use, and create a starting model for the new modelling problem. The goal from that point is to work with experts to revise the starting model to suit the new modelling problem. Discussing specific techniques for revising such a model is out of our scope, but we would note that one technique the team has used in other projects is to begin with a gradual reveal of the full starting model, giving experts a systematic chance to comment on smaller, digestible parts of the model as they are revealed, rather than requiring them to understand and propose revisions to the full (possibly complex) model all at once. 

\underline{Case study:} As noted, our earliest attempts to run workshops with a starting model did not meet with great success, requiring on-the-fly changes to the elicitation approach within those workshops. However, as the models matured, the default approach in later elicitation sessions was to work with a starting model (that is, whichever version of the model was considered the latest for the given modelling problem), particularly for the 1-on-1 sessions. It is important to note that this was not the only option (though it might at first seem to be) --- we {\em could} have started such sessions with an empty model, and then incorporated the elicited causal knowledge back into other existing CKBNs. Indeed, starting with close-to-empty models was the default for the immune family of models, due to the wide variety of opinions we encountered on how the immune system might respond.

\paragraph{Qualitative features} Annotating arcs and relationships with qualitative features (such as the direction of a relationship) is not strictly part of the structure, but this information {\em is} an important part of the causal knowledge base. The typical assumption that holds in a causal BN is that the cause's presence increases the probability of the effect's presence, or that an increase in the strength of a cause probabilistically increases the strength of the effect. That is, $A \rightarrow B$ is by default interpreted to mean that when $A$ is true or present or on (or has increased), then the probability $B$ being true or present or on (or increased) increases.

While this is strictly incorrect --- $A \rightarrow B$, even in a causal BN, merely means that intervening on $A$ will lead to some change in the probability distribution over $B$ --- we nonetheless advocate this as the default interpretation for an arc in a CKBN. Specifically, we recommend nodes represent Boolean propositions that, when true, increase the chance of their (Boolean) effects being true. This is because it is generally a good default when working with experts, providing an intuitive understanding of the structure without the need for careful reading of a definition. To accommodate the very many alternatives, such as causes that {\em decrease} the chance of an effect being true, or enable other causes to operate, etc., arcs and relationships should be annotated to indicate these alternative types of influence. The QGeNIe package provides opportunities for these kinds of annotations \cite{}.

To make this approach work well, it is necessary to reformulate node names or definitions to represent Boolean propositions. For example, while it's perfectly valid in a causal BN to have a node called `Perfusion' with states `Good' and `Poor', we recommend that either `Good Perfusion' or `Poor Perfusion' be specified instead if nothing is lost by treating the node as Boolean. In some cases, only the definition needs to be specified correctly. For example, the states of `Infection' might be considered to be `Present' or `Not Present' --- but we could also just specify that nodes are assumed to be prefixed with either `has' or `is' when the node is otherwise not an obvious proposition. Hence, Infection would be interpreted as 'Has Infection', which acts as a proposition.\footnote{More accurately, it is a predicate, but the subject is often implied in a CKBN, and hence a predicate in a BN always refers to some well-understood proposition --- and has to, if the BN is to be valid.}

It is important to keep in mind that the optimal Boolean form for a node will depend on {\em all} the nodes in the causal BN. For example, if background conditions affect perfusion which in turn leads to symptoms, the best formulation for perfusion is likely to be `Poor Perfusion' rather than `Good Perfusion', as the true states in all the causal BN's nodes coincide with that form. Inevitably, not all nodes will align well, and relationships should be annotated where they have a different interpretation to the default.

\underline{Case study:} In the Respiratory and Complication BNs, nodes were initially assigned no particular Boolean interpretation. Thus, nodes such as intravascular volume, end-organ perfusion, and the various organ function nodes represented the state of those systems or functions, whatever their states may be. However, since experts consistently interpreted relationships as positive, Boolean and causal, we found communication was greatly improved by formulating the node names this way where feasible. It is worth noting that this approach influenced but did not determine the definition and state spaces for the application BN (the Progression BN); for example, `Reduced functional intravascular volume' in the CKBN became `ci\_intravas\_volume\_t{0,1}' (defined as functional intravascular volume), with the states `reduced' and `normal' in the application BN.

\subsection{Checking and revising CKBN structures} \label{sec:checkBn}

In the process of developing the structure of a CKBN, and also once a candidate structure has been developed, there are several checks and common changes that often should be made to ensure the CKBN provides a suitable encoding of the causal knowledge. Many of these techniques are already described as a part of KEBN, as well as other approaches to developing BNs, so we describe them only enough to illustrate how they can support contributions to a causal knowledge base. 

\paragraph{Direct or indirect causes} Causes may be shown as directly affecting a node, indirectly affecting the node via intermediate causes, or both. While a decision to keep or remove an arc always depends on the context, a good heuristic is to ensure the node is connected via fewer paths. In particular, if the contribution to the effect on the node via the direct pathway is likely to be weak, then it can usually be pruned.

A major problem with this approach is that it can lead to longer pathways where omissions of small influences begin to add up (or more accurately, multiply together). In particular, the influence of the original cause becomes increasingly diluted as it passes each node in the chain of causes.\footnote{For example, if, in the chain $A \rightarrow B \rightarrow C$, $A$ accounts for 90\% of the information in $B$, then it can account for {\em at most} 90\% of the information in $C$, and only that if $B$ determines $C$. The longer the chain, the more diluted the influence.} This is not such a concern in a general CKBN, however this dilution effect needs to be accounted for if such models are qualitatively parameterised or used to build application BNs.

\underline{Case study:} This technique was used consistently in all the workshops while eliciting a CKBN. As the CKBN expanded in size, experts were regularly asked to decide whether just the direct, just the indirect, or both pathways were required.
This frequently led to the omission of one or the other of the pathways.

\paragraph{Many parents or many children} An approach related to checking for direct and indirect causes is checking the number of parents a node may have. It may seem simplistic, but simply checking nodes that have the largest number of parents (particularly, diverse parents) can produce substantial improvements to the way causal knowledge is recorded in a CKBN. This is because a large number of diverse parents is a strong indication of a weak understanding of the causal mechanisms involved. Intermediate nodes in such cases can be introduced in two ways in such cases: either as logical (noisily deterministic) intermediate nodes, that do not play a distinct causal role; or intermediate nodes that do play a distinct causal role. The former is useful for simplifying the CKBN and particularly grouping like events together, while the latter is of course useful for clarifying the causal story.

Checking the number of children a node has may be less fruitful, but can still be useful. For example, age is often a variable that affects many things, and this is not necessarily much of a problem; at the same time, if age can be made to influence fewer variables directly, it suggests the representation of the mechanisms is much more powerful.\footnote{Age affects everything, but causes nothing. This is easy to check: whenever age is posited as a cause, it can easily be replaced by a mechanistic cause.}

\underline{Case study:} In the Complications BN, organ dysfunctions and failures were originally treated as combined nodes. However, the number of parents entering these combined nodes was significant and the story seemed abbreviated and inadequate for capturing the differing ways in which they could arise (as well as the types of management that could be performed). Hence, organ failures were separated out from dysfunctions, taking some (but not all) of their parents with them.

\paragraph{Arc direction} Checking the arc direction is a must in any causal BN. At first, this may seem to be a simple case of ensuring the arc direction is correct. However, any practical and useful CKBN abstracts away {\em many} things, including specific occurrences of an event, specific times and durations between events, and categories and types of events. In many cases, both directions are correct {\em given} the abstractions, so a choice has to be made.\footnote{A choice has to be made even if the CKBN is dynamic, since other parts of the structure will still depend on the choice.} There may be many factors affecting the choice in such cases, however a reasonable heuristic is to choose the direction that represents the most important direction of influence for reaching the outcomes of interest. Such arcs should be annotated to indicate that the cause can also head in the opposite direction if that direction of causal flow is significant; if it is not significant, the annotation can be omitted (or noted as not significant).

\underline{Case study:} An example in which the arc direction could go in either direction is the connection between reduced cardiac output and hypoxia. The chosen direction (reduced cardiac output to hypoxia) contributes much more to the evolution of better or worse outcomes from COVID-19 than its reverse, and hence is the direction used in the CKBN. (Of note, the reverse direction is still partly represented here by the more general relationship in which hypoxia can lead to multi-organ failure.)

\paragraph{Variable definitions and dictionaries} As noted Section~\ref{sec:funcStructCkb}, a CKBN is considered to consist of both a causal structure and a dictionary. When a new CKBN is still being developed and conceptualised, there may not yet be a strong need for a dictionary; in its place, there is essentially a proto-dictionary that consists just of the node titles and any node or arc specific annotations that may have been made. This is entirely appropriate, as {\em too much} time spent on precise definitions that are often already well understood even if not precise (for example, infection at a site, or O2 saturation) can stall development of a CKBN. However, there is a trade-off: in some cases, progress cannot be made without clarifying a definition first. And in all cases, a CKBN that is communicated formally (for example, via a report or publication) should have a  dictionary with clear definitions for all variables.

Variable definitions are not just for communication. They also help to ensure the logic of the CKBN is correct, and that causal knowledge has been captured accurately from its source. A CKBN variable definition has to stand on its own and represent an event that has a clearly defined causal role. The process of checking and providing a definition helps to identify whether the causal role is clearly defined or needs to be divided into multiple variables, or else suggests other changes to the CKBN structure are needed. Importantly, CKBN variables should not directly adopt definitions from elsewhere without proper checking and revision, since those definitions may not align well with the intended causal role. This is especially true for precise definitions in data dictionaries.\footnote{This is because data consists of measurements, and measurements are downstream effects of an event that act as indicators of the original event \cite{}. They very rarely map precisely to the event and frequently do not map to the event in the same way across different datasets, even when standardised.}

\underline{Case study:} For each of the CKBNs, variable dictionaries were created once the structure of the CKBN appeared sufficiently stable. The work involved in creating the dictionaries (particularly in identifying the relevant literature) was non-trivial, hence dictionary updates trailed behind updates to the CKBN structure, and were synchronised and finalised for published CKBNs. During expert elicitation of the models, definitions were considered to be well understood for well known concepts, but experts were regularly prompted to see if the both the definitions and structures appeared reasonable to them (giving experts a chance to check if the definitions fit the causal roles of the variables in the CKBN structure).

\paragraph{Divide and merge} In addition to the broader check of variable definitions above, nodes can be explicitly checked to see if it would be more appropriate to divide them into multiple nodes, or to merge nodes together. The decision is not always a simple matter of matching the true causal story. A simple example of this is two alternative representations of multiple diseases: 1) each disease is represented with its own node, and 2) all diseases are represented in a single node. If the diseases can co-occur, representation (2) would be technically incorrect and could lead to invalid inferences. However, if the probability of co-occurrence is sufficiently low, the error may be of little consequence compared to the benefits of representing them in a single node; and if not, such merges can be made technically correct by specifying an importance ranking over the diseases (e.g., $A$ is true if disease $A$ is present, $B$ is true if disease $A$ is not present, but disease $B$ is, etc.). This works much like the importance ranking used in a Noisy MAX \cite{}.

\underline{Case study:} There were many examples in the CKBNs of choices around the best way to group or divide nodes. For example, in the Respiratory and Complications BNs, systemic immune response and systemic inflammation at various sites could be represented either as a single joint node, or as two separate nodes. Since systemic inflammation is a form of immune response, and there was no notable impact of inflammation independent of other aspects of immune response, it was decided that the two would be merged. By contrast, alveolar epithelial infection (that is, infection of the cells lining the air sacs of the lung) and alveolar endothelial infection (that is, infection of the cells lining the blood vessels passing the air sacs in the lungs) were separated in the Respiratory BN due to their different causal roles in that model, but merged (along with all other respiratory infections) into a general Respiratory infection node in the Complications BN.

\paragraph{Qualitative features} While this has already been discussed in Section~\ref{sec:devCausalBnStructure}, it is worth noting that this technique is also of particular use in checking the structure of a CKBN. In particular, auditing the nodes of a CKBN and attempting to identify the logical or other qualitative relationships of the parents to the child (whether or not that knowledge needs to be captured) is useful in identifying inconsistencies or confusions around the definitions in the causal structure.

\underline{Case study:} As noted, the logical relationship between V/Q mismatch and Shunt was discussed after the Respiratory BN had been elicited, and this clarified the definitions, causal story and intervention possibilities around these nodes. A further example was around the logical relationships between Hypercapnia, Hypoxaemia, Reduced lung compliance and Dyspnoea, leading to introduction of the Perceived need for air node to better represent the causal mechanisms.

\paragraph{Qualitative parameterisation} While discussed earlier, qualitative parameterisation is invaluable in ensuring that the CKBNs are capable of fulfilling their purpose and that they can support future applied models. A qualitative parameterisation can be performed without any reference to specific data or high quality parameter estimates. Instead, the qualitative parameterisation can be done based on the understanding that the modellers have of the system gathered through the knowledge collection process up to that point, supplemented by support from experts and the literature on approximate numbers, directions and relationships where there are gaps that would make completing the parameterisation too difficult.

This technique has several benefits: it ensures that CKBN structures are internally consistent; it ensures that structures and qualitative causal features match the knowledge of experts, literature and other causal knowledge sources; it identifies gaps in the current knowledge base; it can identify misunderstandings and particularly definitions that are too loose to work well; it can also identify aspects of the CKBN that may be more or less important, and help guide where to focus efforts in revising the CKBNs. Beyond the direct support to the causal knowledge base, qualitative parameterisation can also support quantitative work, including sensitivity analyses, parameter training (in the form of priors), and validation of applied BNs.

\underline{Case study:} All of the CKBNs that were published, and many that were not, were qualitatively parameterised. The process was invaluable in identifying how robust the models were, where further work was required, and also in communicating the behaviour and intent of the CKBNs to experts and others beyond. The Testing BN made particularly good use of a (higher quality) qualitative parameterisation to communicate and illustrate the much richer causal story that determines most sensitivity and specificity estimates.